\documentclass{article}
\usepackage{newunicodechar}
\newunicodechar{≈}{\approx}

\usepackage[authoryear, sort&compress, round]{natbib}
\usepackage[preprint,logo]{agibot_enerverse} 

\usepackage[utf8]{inputenc} 
\usepackage[T1]{fontenc}    
\usepackage{hyperref}       
\usepackage{url}            
\usepackage{booktabs}       
\usepackage{amsfonts}       
\usepackage{nicefrac}       
\usepackage{microtype}      
\usepackage{xcolor}   
\usepackage{amsmath}
\usepackage{graphicx}
\usepackage{caption}
\usepackage{wrapfig}
\usepackage{pifont}
\usepackage{colortbl} 
\usepackage{multirow} 
\usepackage[table]{xcolor}
\usepackage[absolute,overlay]{textpos}

\definecolor{green}{RGB}{0,150,10}
\definecolor{blue}{RGB}{0,148,181}
\definecolor{orange}{RGB}{194,153,107}

\definecolor{grey}{HTML}{999999}
\definecolor{lightblue}{HTML}{B0C4DE}
\definecolor{purple}{HTML}{E3BBED}
\definecolor{orange}{HTML}{ffdab9}
\definecolor{cadetblue}{HTML}{5F9EA0}
\definecolor{darksalmon}{rgb}{0.91, 0.59, 0.48}
\definecolor{forestgreen}{rgb}{0.13, 0.55, 0.13}
\definecolor{BlueGreen}{rgb}{0.0, 0.55, 0.55}
\definecolor{RedOrange}{rgb}{1.0, 0.27, 0.0}

\usepackage{xspace}
\newcommand{\SIM}{\textsc{world simulator}\xspace} 

\title{ \raisebox{-0.005\textheight}{\includegraphics[width=0.055\textwidth]{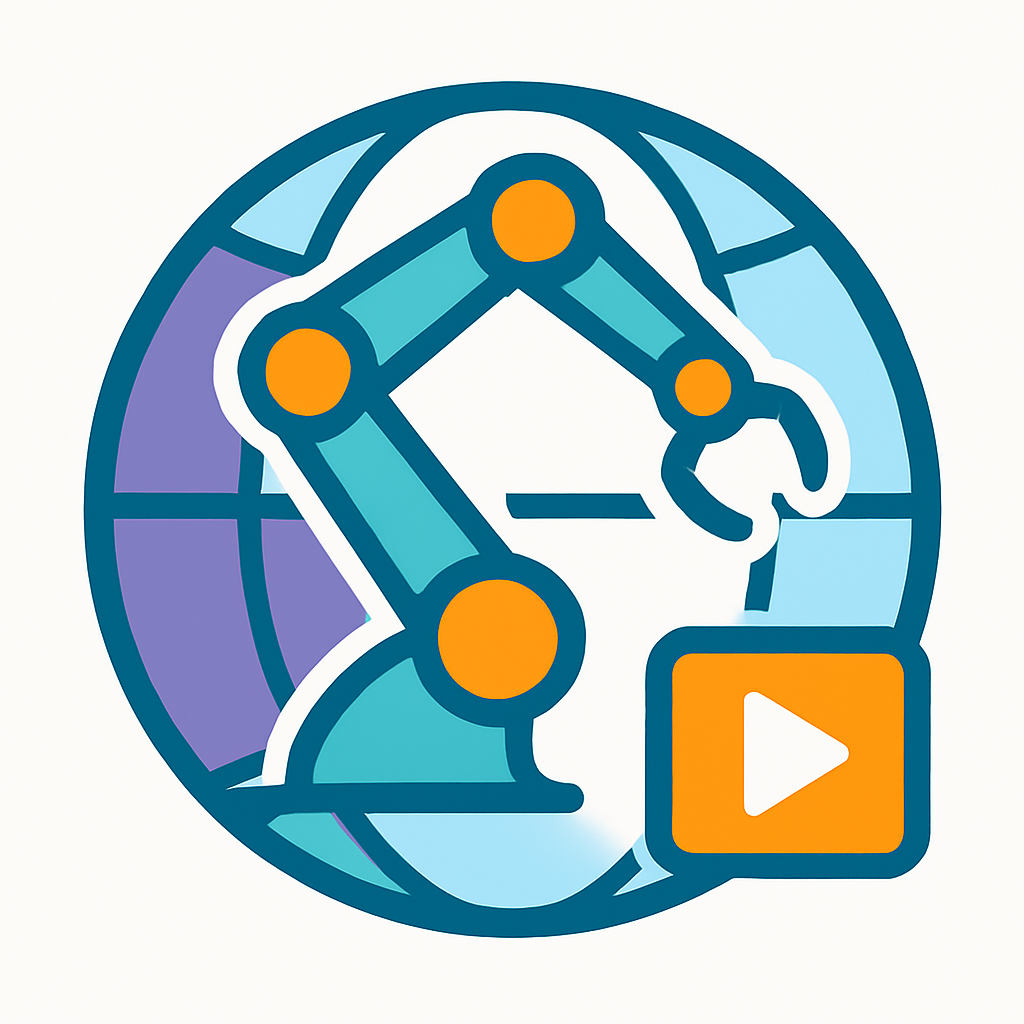}}
 Genie Envisioner: A Unified World Foundation Platform for Robotic Manipulation}

\author{%
 Yue Liao$^{*}$ \quad
   Pengfei Zhou$^{*}$ \quad Siyuan Huang$^{*}$ \quad Donglin Yang \quad Shengcong Chen
  \\
  \textbf{ Yuxin Jiang \quad Yue Hu\quad  Jingbin Cai \quad Si Liu \quad Jianlan Luo \quad Liliang Chen$^{\dagger}$}\\  \textbf{ Shuicheng Yan$^{\diamond}$ \quad Maoqing Yao$^{\diamond}$ \quad Guanghui Ren$^{\dagger\diamond}$}\\[0.5em]
   AgiBot Genie Team \quad LV-NUS Lab \quad
  BUAA \\[0.5em]
\normalsize ~\url{https://genie-envisioner.github.io}
}

\begin{document}

\maketitle
\let\tempfootnote\thefootnote
\let\thefootnote\relax
\footnotetext{$^*$ Equal Contribution. $\dagger$ Project Leader. $^\diamond$ Corresponding Author. }

\begin{figure}[ht]
    \begin{center}
\centerline{\includegraphics[width=1\linewidth]{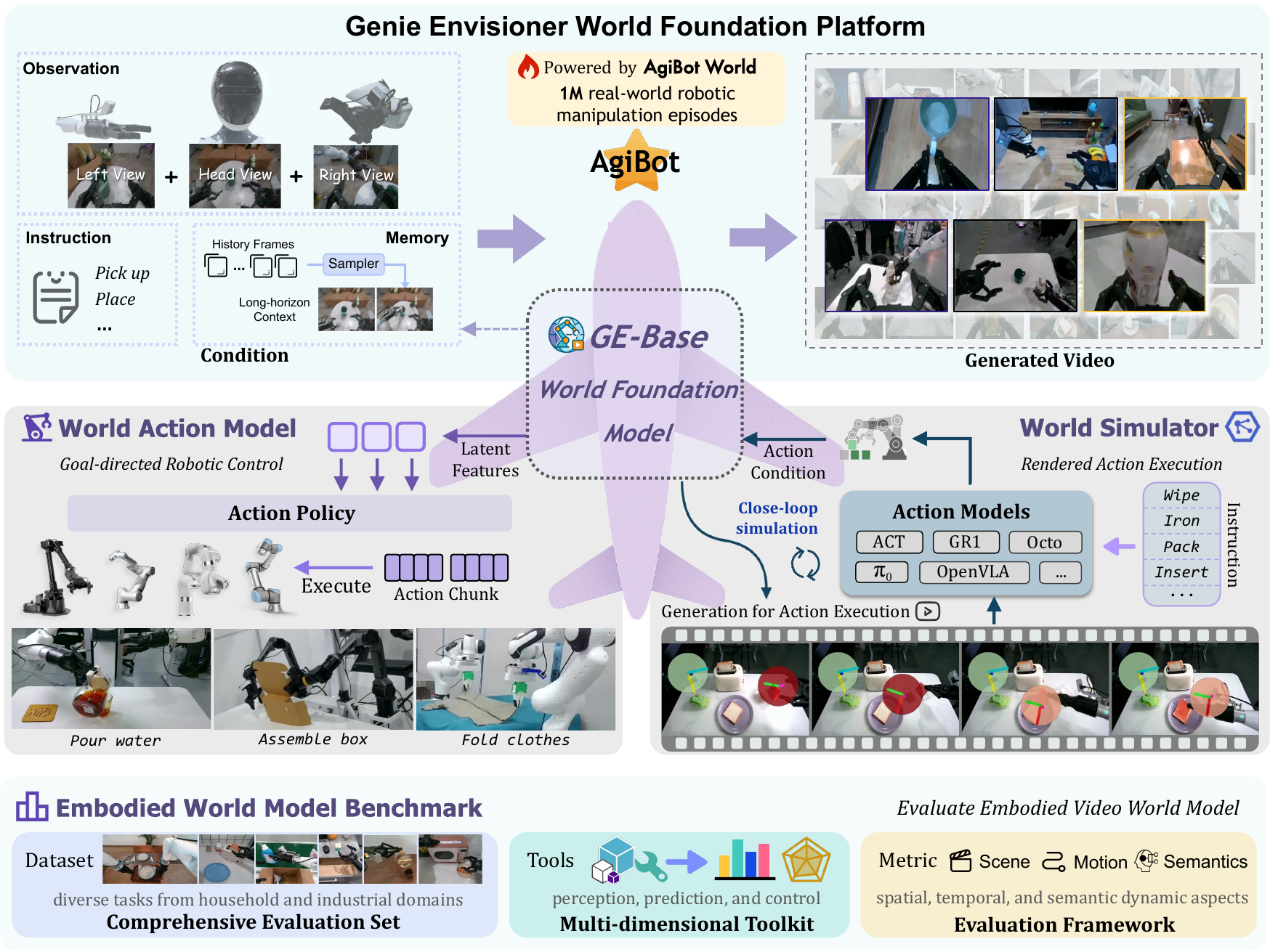}}

        \caption{\textbf{Overview of the Genie Envisioner World Foundation Platform.}
Genie Envisioner is a unified world foundation platform that integrates manipulation policy learning and evaluation within a single video-generative framework. At its core lies GE-Base, a large-scale world model that encodes the spatial, temporal, and semantic structure of robotic interactions. Built around it are two key functional modules: GE-Act, a world action model that infers instruction-conditioned policies, and GE-Sim, a video-based \SIM that enables closed-loop execution through action-conditioned generation. The platform is complemented by EWMBench, an integrated evaluation suite that assesses visual fidelity, physical plausibility, and instruction-policy alignment. GE thus provides a practical and scalable foundation for general intelligence embodiment.}
        \label{fig:banner}
    \end{center}
\end{figure}

\begin{abstract}

We introduce Genie Envisioner (GE), a unified world foundation platform for robotic manipulation that integrates policy learning, evaluation, and simulation within a single video-generative framework. At its core, GE-Base is a large-scale, instruction-conditioned video diffusion model that captures the spatial, temporal, and semantic dynamics of real-world robotic interactions in a structured latent space. Built upon this foundation, GE-Act maps latent representations to executable action trajectories through a lightweight, flow-matching decoder, enabling precise and generalizable policy inference across diverse embodiments with minimal supervision. To support scalable evaluation and training, GE-Sim serves as an action-conditioned neural simulator, producing high-fidelity rollouts for closed-loop policy development. The platform is further equipped with EWMBench, a standardized benchmark suite measuring visual fidelity, physical consistency, and instruction-action alignment. Together, these components establish Genie Envisioner as a scalable and practical foundation for instruction-driven, general-purpose embodied intelligence. All code, models, and benchmarks will be released publicly.
\end{abstract}

\begin{figure}
\vspace{-0.7cm}
    \begin{center}
    \centerline{\includegraphics[width=.99\linewidth]{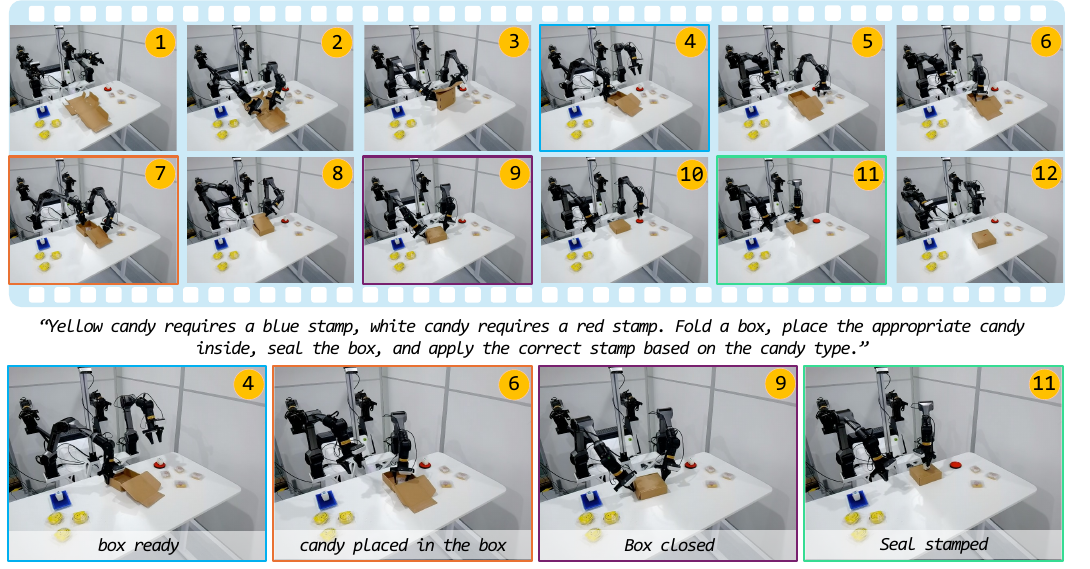}}

        \caption{\textbf{ Real-world demonstration of GE-Act on a novel robot embodiment, Agilex Cobot Magic, unseen during pretraining.}  With only one hour of embodiment- and task-specific teleoperation data for post-training, GE-Act successfully executes a complex manipulation task involving fine-grained control of deformable objects and memory-based decision making.  
        Given a general packaging rule, the robot is required to complete the packing process for each item accordingly. Here, we showcase the detailed execution of the first packing cycle.
        The robot first stacks a deformable box, places a target object inside based on instruction, and closes the lid, \emph{rendering the object no longer visible}. It then correctly selects and applies the appropriate stamp, matching the object type, relying solely on internal memory.  This showcases GE’s generalization to new embodiments, its precise handling of deformable materials, and its ability to retain task-relevant memory across steps.
        .}
        \label{fig:Ge-cap}
    \end{center}
\end{figure}

\section{Introduction}
\vspace{0.5em}
\begin{quote}
\textit{"The best way to predict the future is to invent it."}
\hfill — Alan Kay
\end{quote}
\vspace{0.5em}

Embodied agents that sense, reason, and act in the physical world represent the next frontier of AI systems. 
At its core, a fundamental research challenge remains: developing scalable and robust robotic manipulation capabilities - the ability to purposefully interact with and control the physical environment through selective contact~\citep{mason-manipulation}. 
Despite considerable progress has been made in this domain, ranging from analytic~\citep{berenson2009manipulation,stilman2007task}, model-based frameworks~\citep{ebert2018visual,janner2019trust,nagabandi2020deep} to data-driven approaches that learn manipulation policies from large-scale datasets~\citep{kim2024openvla,black2024pi_0,brohan2023rt2,bu2025univla}, existing systems typically rely on a \emph{patchwork} of separate data‑collection, training, and evaluation stages. Each stage demands bespoke infrastructure, manual curation, and task‑specific tuning; the resulting friction could potentially slow down iteration, obscure failure modes, and impede reproducibility at scale. These fragmented stages underscore the absence of an integrated framework capable of learning and evaluating manipulation policies in a unified manner.

To this end, we introduce Genie Envisioner (GE), a unified platform that collapses robot sensing, policy learning, and evaluation into a single closed-loop video generative world model, as illustrated in Figure~\ref{fig:banner}. At its core lies \textbf{GE‑Base}, an instruction-conditioned, multi-view video diffusion model trained on approximately 3,000 hours of video-language paired data spanning over one million real-world robotic manipulation episodes from the AgiBot‑World‑Beta dataset~\citep{bu2025agibot}. Conditioned on robot’s visual observations, GE‑Base autoregressively generates video chunks that capture the temporal evolution of manipulation behaviors following high-level instructions. 
Leveraging robotic domain adaptation pretraining, 
GE‑Base establishes a mapping from language instructions to an embodied visual space, capturing the essence of robotic manipulation by modeling the spatial, temporal, and semantic regularities of real-world interactions. It achieves this by inferring latent trajectories that jointly encode the robot’s perceptual inputs and the anticipated evolution of the scene under plausible action sequences.
To bridge the gap between visual representations and executable robotic control, we introduce \textbf{GE‑Act}, a lightweight parallel flow-matching action model. GE‑Act translates visual latent features, conditioned on language instructions, into fine-grained and low-latency motor commands, enabling direct and efficient mapping from perception and instruction to executable physical actions.
Beyond policy learning, simulation plays a critical role in enabling scalable training, safety validation, and fast iteration for robotic systems. To this end, we introduce \textbf{GE-Sim}, which leverages the embodied video generation capabilities of GE‑Base and repurposes its generative dynamics into an action-conditioned \SIM. GE-Sim supports closed-loop policy evaluation through video-based simulation, achieving speeds significantly faster than real-world execution. After designing the core foundation model, a critical challenge remains: evaluating whether the generated videos faithfully simulate robotic behaviors. This requires moving beyond generic perceptual metrics to assess whether the synthesized behaviors are both physically grounded and semantically aligned with the given instructions.
To address this, we propose the Embodied World Model Benchmark (\textbf{EWMBench}), a principled evaluation suite that directly benchmarks video generative neural world simulators in terms of visual fidelity, physical consistency, and instruction–action alignment.  Therefore, GE constructs a unified video-based robotic vision space that facilitates the learning, simulation, and evaluation of action policies within a perceptually grounded framework.  Different from mainstream vision-language-action (VLA) methods~\citep{kim2024openvla,black2024pi_0} that rely on vision-language models (VLMs)~\citep{bai2025qwen2,chen2024internvl,abouelenin2025phi} to map visual inputs into a semantic linguistic space and learn action policies from this language-centric representation, GE constructs a vision-centric space through generative video modeling. This space preserves detailed spatial and temporal cues, enabling more faithful modeling of robot–environment dynamics and supporting end-to-end policy learning and evaluation within a single, coherent platform.

To comprehensively evaluate the capabilities of GE across embodied video generation, policy learning, and simulation, we conduct extensive experiments on a diverse set of real-world robotic manipulation tasks. GE-Act achieves low-latency end-to-end control by generating \emph{54-step torque trajectories within 200 ms on a commodity GPU}. It delivers precise task execution on the in-domain AgiBot G1 platform and demonstrates strong cross-embodiment generalization to novel systems, including Dual Franka and Agilex Cobot Magic, with only \emph{1 hour of teleoperated demonstrations}, outperforming task-specific baselines~\citep{black2024pi_0,bjorck2025gr00t,bu2025univla}. GE-Act proves effective across a wide range of scenarios and tasks, including industrial applications such as conveyor-based moving object manipulation, and household tasks such as cooking, table cleaning, and pouring. Beyond these standard manipulation tasks, GE-Act's visual world modeling enables it to handle long-horizon, memory-intensive sequences, as shown in Figure~\ref{fig:Ge-cap}.
Furthermore, GE-Sim enables policy rollout evaluation at thousands of episodes per hour via distributed cluster parallelization, substantially accelerating the assessment of manipulation capabilities and policy training. EWMBench provides a comprehensive evaluation framework for video-based world models, systematically benchmarking GE-Base against state-of-the-art video generation models. The results reveal superior performance of GE-Base in robotic world modeling, with strong alignment to human assessments, underscoring its role as a foundational component of the unified GE platform.

Together, these contributions position Genie Envisioner as a practical, scalable foundation for real‑world manipulation, facilitating downstream research. All code, pretrained models, and the complete EWMBench suite will be open‑sourced upon publication to accelerate future research.

\begin{figure}[t]
    \begin{center}
\centerline{\includegraphics[width=1\linewidth]{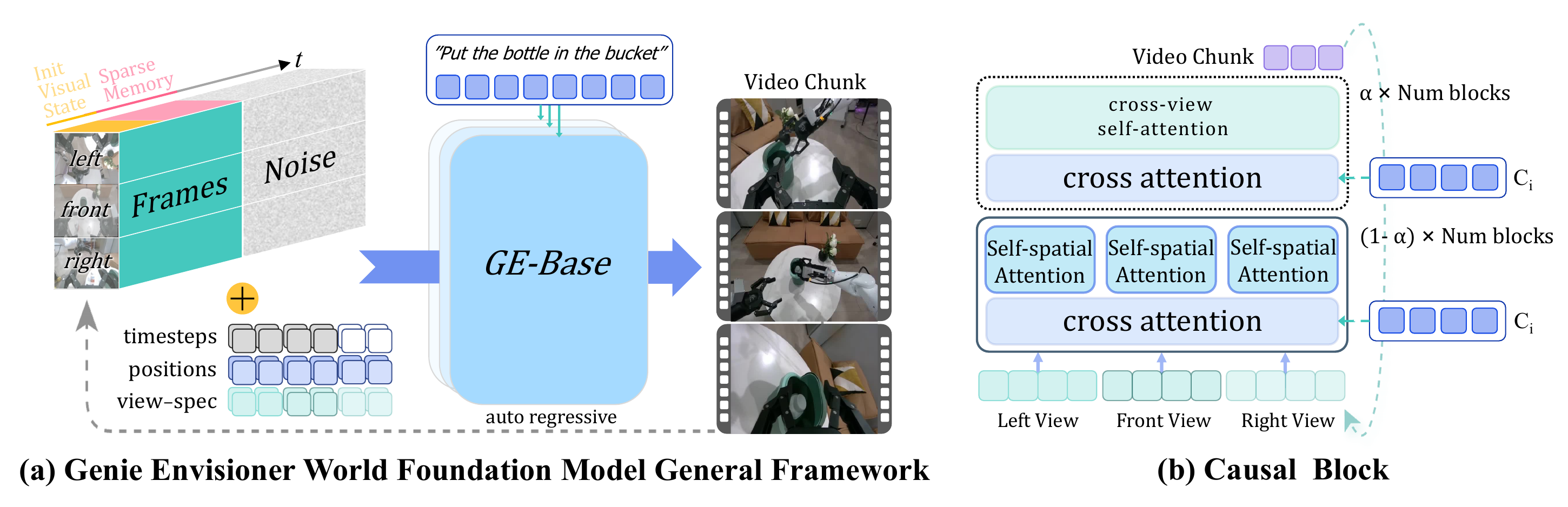}}
        \caption{\textbf{Overview of the GE-Base World Foundation Model.} (a) An illustration of the autoregressive video generation process. Given multi-view visual conditions, including the initial observation and sparse memory, along with corresponding noise and positional embeddings, the model generates the next multi-view video chunk conditioned on a language instruction.
(b) A dedicated causal block facilitates information exchange across different views, ensuring spatial consistency during multi-view video chunk generation.}
        \label{fig:framework-ge-base}
    \end{center}
\end{figure}

\section{GE-Base: World Foundation Model}

In this section, we present \textbf{GE-Base}, the core component of Genie-Envisioner.
Our objective is to extend the predictive capacity of general video generation models toward constructing an \emph{embodied predictive representation}—a unified generative formulation that anticipates future robot-environment interactions conditioned on task instructions and grounded in the agent’s physical embodiment.
To this end, we formulate robotic video world modeling as a text-and-image-to-video generation problem: given a language instruction and an initial visual observation, the model forecasts future video segments that reflect plausible and coherent robotic behaviors.
A key design feature of GE‑Base is its sparse memory mechanism, which augments the current visual input with long-term historical context, enabling stronger temporal reasoning through a unified visual condition.
Built upon this formulation, GE‑Base adopts a video diffusion transformer architecture and incorporates a robotic-adaptive pretraining strategy that transfers knowledge from generic video datasets into the embodied robotic domain.
We demonstrate the effectiveness of GE‑Base on real-world robotic manipulation video generation. Experimental results show that GE‑Base produces instruction-aligned, temporally coherent video sequences that generalize well across diverse manipulation tasks and embodiments.
\subsection{Basic Architecture}
To align with the sequential nature of robotic manipulation data, we adopt an autoregressive video generation framework that segments the output into discrete video chunks, each containing $N$ frames.
At each step $t$, the world foundation model $\mathcal{W}$ generates the next chunk $\mathbf{x}_{1:N}^{(t)}$
conditioned on three components: the initial visual observation $\mathbf{x}_0$, the language instruction $q$, and a sparse memory $\hat{\mathbf{x}}_{0:{t-1}}$, constructed by sparsely sampling long-term historical frames from previous steps.
The generation process is formally defined as:

$$
\mathbf{x}_{1:N}^{(t)} = \mathcal{W}(\hat{\mathbf{x}}_{0:t-1},\ \mathbf{x}_0,\ q).
$$

This formulation enables the progressive generation of temporally coherent video segments, grounded in both visual and instruction conditions. By integrating long-term sparse memory into the visual state, rather than relying solely on recent frames, the model effectively captures extended temporal dependencies while maintaining semantic alignment and visual consistency throughout the manipulation process.

To balance efficiency and capacity in robotic video modeling, we adopt a compact video generation model as the core architecture. Our GE-Base world model $\mathcal{W}$ is designed with flexibility in mind, allowing seamless integration with various diffusion transformer (DiT)-based video generation models. Specifically, we select LTX-Video 2B~\citep{HaCohen2024LTXVideo} and COSMOS2 2B~\citep{agarwal2025cosmos} as our base models. LTX-Video provides a faster and more lightweight architecture, supporting efficient downstream action policy prediction, whereas COSMOS2 offers higher-quality video synthesis, making it well-suited for high-fidelity simulation. Considering the egocentric nature of perception in dual-arm robotic systems, we extend $\mathcal{W}$ into a multi-view, language-and-image-conditioned generation framework that leverages temporally synchronized inputs from three onboard cameras: a head-mounted view ($v^h$) and two wrist-mounted views ($v^l$, $v^r$). Each frame in $x_0$, $\hat{\mathbf{x}}_{0:t-1}$, and $x_t$ follows this tri-view observation structure.

As illustrated in Figure~\ref{fig:framework-ge-base}, the generation pipeline begins by encoding multi-view observations from the initial visual observation $x_0$ and the sparse memory $\hat{\mathbf{x}}_{0:t-1}$ using a shared video encoder $\mathcal{E}$. For each view, we obtain latent visual tokens, denoted as $\mathcal{E}(v_0^{(i)})$ and $\mathcal{E}(v_{t-1}^{(i)})$ for $i \in {h, l, r}$. The visual token sequence for each view is composed by concatenating tokens from $x_0$ and $\hat{\mathbf{x}}_{0:t-1}$. Corresponding to each view, a distinct noise map $z^{(i)}$ is initialized to guide generation. To preserve spatiotemporal alignment while distinguishing viewpoint-specific information, we augment each token and noise input with both a 2D rotary positional embedding $e_{\text{pos}}$ and a view-specific learnable embedding $e_{\text{view}}$. These enriched tokens and noise maps from all views are concatenated and further embedded with a timestep encoding $e_t$, then fed into the DiT backbone for autoregressive generation of the next video chunk.

To facilitate coherent reasoning across multiple views, we extend standard spatial self-attention over $(H, W)$ to cross-view self-attention over $(N, H, W)$, where $N$ denotes the number of camera perspectives. Hidden states are reshaped to $(B, N, T, H, W, C)$ to enable joint cross-view reasoning. To ensure computational tractability, cross-view attention is sparsely inserted into selected DiT blocks, while the remaining blocks treat views independently by folding the $N$ dimension into the batch dimension, yielding a shape of $(B \cdot N, T, H, W, C)$. This hybrid attention scheme balances view-level consistency and efficiency.

To incorporate semantic task-level guidance, the instruction $q$ is processed using a frozen T5-XXL encoder~\citep{raffel2020exploring}, producing a set of text embeddings $\mathcal{T}(q)$. These are integrated into the visual token stream through cross-attention layers within the DiT, allowing the model to align video generation with the instruction semantics.

Given this design, the world model $\mathcal{W}$ predicts the next video chunk $\hat{x}_t$ as:

$$
\hat{x}_t = \mathcal{W} \left( \{v_0^{(i)}, v_{\hat{t}}^{(i)}, z^{(i)}\}_{i \in \{h, l, r\}},\; \mathcal{T}(q) \right),
$$ 

where $v_0^{(i)}$ and $v_{\hat{t}}^{(i)}$ denote the encoded initial and historical visual tokens from view $i$, $z^{(i)}$ represents the corresponding view-specific noise map, and $\mathcal{T}(q)$ is the encoded language instruction.

This unified modeling paradigm enables $\mathcal{W}$ to jointly capture spatial layouts, temporal dynamics, and semantic intent—yielding coherent and controllable predictions of embodied robotic behavior.

\subsection{World Model Pre-training}
A core challenge in building video-based world models for robotic manipulation lies in adapting general video generation capabilities to the structured dynamics and semantics of the embodied robotic domain. To address this, we develop a multi-stage pre-training framework that progressively aligns the model’s spatiotemporal representations with the distributional characteristics of real-world robot behavior. This section outlines our data curation pipeline and the corresponding training strategies for domain adaptation. During training, sparse memory frames are randomly sampled from preceding video history, serving as a form of data augmentation. This design increases the difficulty of future prediction and enhances the model’s robustness to temporal variation, ultimately improving its generalization to diverse manipulation scenarios.

\begin{figure}[t]
    \begin{center}
\centerline{\includegraphics[width=1\linewidth]{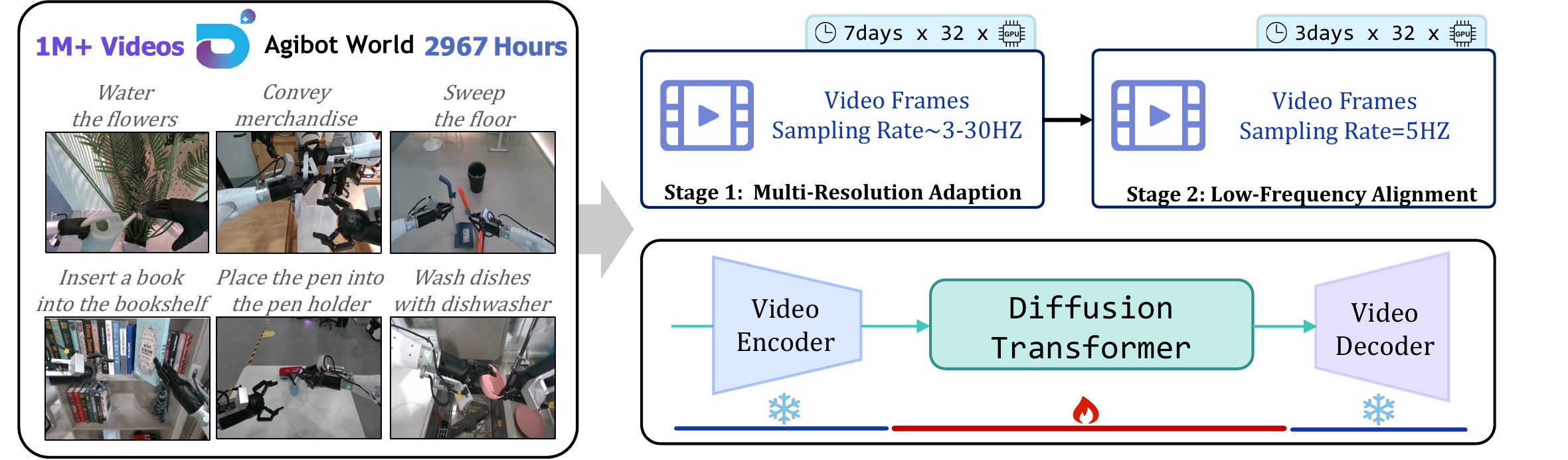}}
        \caption{\textbf{Overview of the GE-Base Training Process.}
GE-Base is pre-trained on AgiBot-World-Beta, a large-scale real-world dual-arm robotic manipulation dataset containing 1 million instruction-aligned, multi-view video sequences. The training begins with a domain adaptation phase, transferring general video generation capabilities into the robotic domain using high-frame-rate sequences and mixed sampling strategies to enhance robustness. This is followed by a low-frame-rate fine-tuning stage designed to align the model with the temporal resolution required for downstream action policy training. Throughout the process, the video encoder and video decoder remain fixed.}
        \label{fig:train-video}
    \end{center}
\end{figure}

\noindent\textbf{Data Curation.}
We adopt the AgiBot-World-Beta~\citep{bu2025agibot} dataset as the foundation for pretraining. This dataset comprises approximately one million high-quality real-world dual-arm robotic manipulation episodes, totaling $2,967$ hours, collected via human teleoperation. The dataset spans a diverse range of tasks, object categories, and environments, with each trajectory annotated with natural language instructions, multi-view visual observations, and structured action policies. To adapt the dataset for video-based modeling, we extract temporally synchronized video streams from three calibrated camera viewpoints and ensure semantic consistency between each video segment and its paired instruction. This preprocessing step results in high-quality text–video pairs that reflect coherent and executable manipulation behavior. To accommodate different learning objectives across pretraining stages, we employ variable frame sampling strategies that balance temporal resolution and training stability.

\begin{figure}[t]
    \begin{center}
\centerline{\includegraphics[width=1\linewidth]{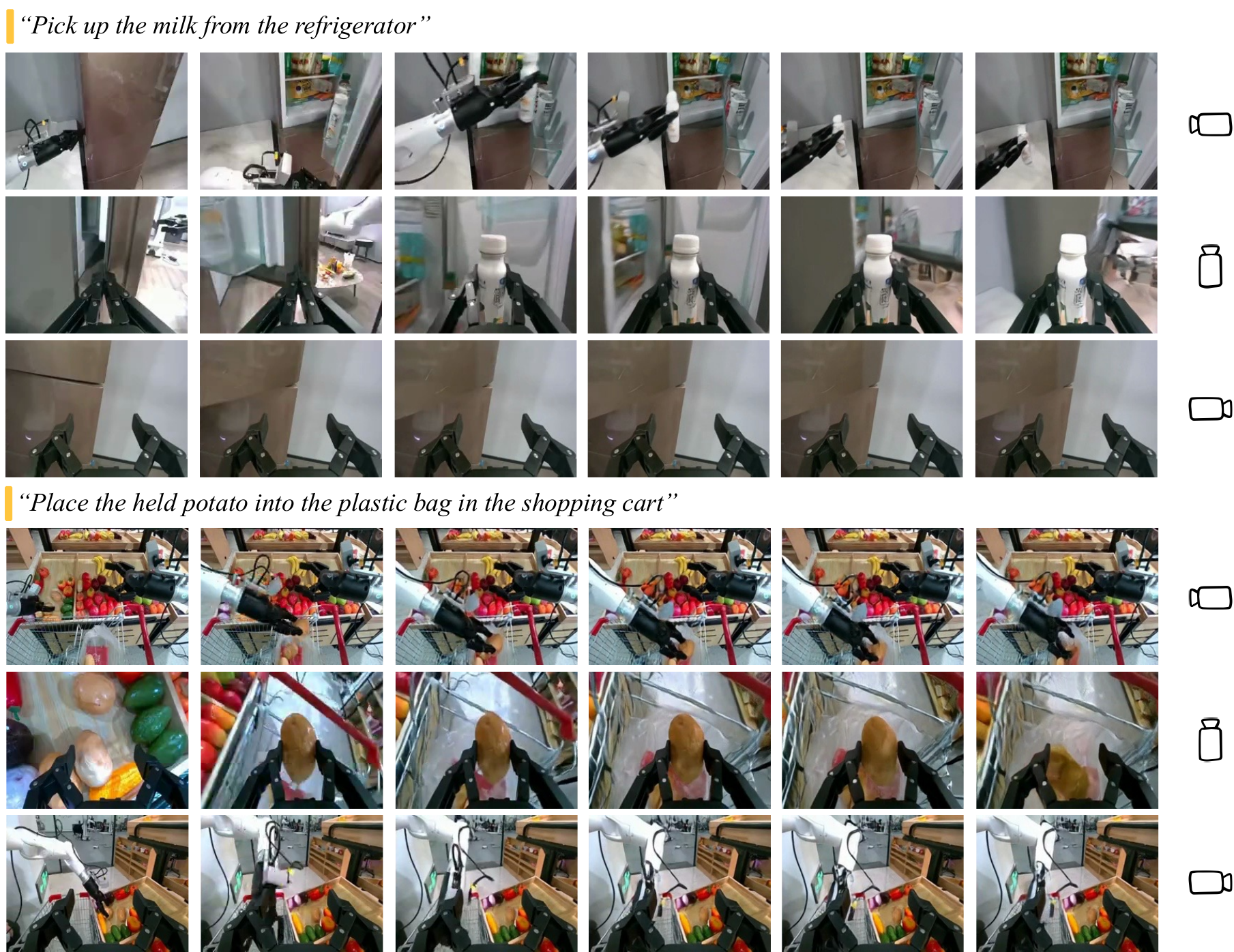}}
        \caption{\textbf{Multi-View Robotic Manipulation Videos Generated on AgiBot G1 by GE-Base. }
We visualize robotic manipulation sequences generated by GE-Base across two tasks involving varied objects and environments. For each example, videos from three views are presented, \emph{i.e.}, the head-mounted, left-, and right-arm cameras, respectively.}
        \label{fig:demo-video-gen}
    \end{center}
\end{figure}

\noindent\textbf{Stage I: Multi-Resolution Temporal Adaptation (GE-Base-MR).}
The first stage aims to bridge the gap between generic video representation learning and robotic-specific motion dynamics. We pretrain the model on 57-frame video sequences randomly sampled at frame rates between 3 Hz and 30 Hz. Each training sample includes four sparse memory frames, randomly drawn from prior video history to enhance temporal diversity. These clips are encoded into an 8-frame latent space via a pretrained VAE, where noise is added and the model is optimized through a denoising objective.

This training setup exposes the model, referred to as GE-Base-MR, to a wide spectrum of motion speeds and temporal patterns, encouraging it to learn spatiotemporal representations invariant to sampling rates. Conditioning on both visual observations and language instructions, the model learns to map high-level task intent to low-level visual dynamics while maintaining robustness to partial observations. This design is essential for real-world deployment, where sensor latency, frame drops, and asynchronous data are common. After this stage, GE-Base-MR is capable of generating high-quality robotic manipulation videos that accurately capture motion dynamics and visual consistency. The model is trained end-to-end on the AgiBot-World-Beta dataset using 32 NVIDIA A100 GPUs for approximately seven days.

\noindent\textbf{Stage II: Low-Frequency Policy Alignment (GE-Base-LF).}
To improve training efficiency and better align with the temporal abstraction used in downstream action modeling, we fine-tune GE-Base-MR using low-frame-rate video sequences. Specifically, we sample 9-frame clips at a fixed rate of 5 Hz and provide 4 additional sparse memory frames as temporal context. These sequences are mapped into a compact latent space consisting of two latent frames via a pretrained video encoder, whose parameters remain frozen. Only the video generation components are updated during this phase. The resulting model, GE-Base-LF, is optimized to capture semantically meaningful transitions under sparse visual sampling. Training remains end-to-end for the generative pathway and is conditioned on both task instructions and visual conditions. This process effectively aligns the video DiT with the temporal abstraction used in control, enabling reliable video feedback at the granularity of discrete action steps. GE-Base-LF serves as a critical foundation for subsequent action model pretraining and is trained for approximately three days using 32 NVIDIA A100 GPUs.

\subsection{ Robotic Manipulation Video Generation via GE-Base}
We generate dual-arm robotic manipulation videos using GE-Base, leveraging the LTX-Video 2B architecture (with additional base architectures currently being explored). This process follows an autoregressive approach, where each step generates a new video chunk conditioned on the initial observation, a series of memory frames, and the language instruction. The generation proceeds iteratively until the task specified by the instruction is fully executed, resulting in a seamless video sequence that precisely captures the complete manipulation procedure.

At inference time, memory frames are uniformly sampled from prior video chunks at fixed intervals, ensuring stable temporal dynamics and consistent prediction. We evaluate this pipeline on real-world dual-arm robotic manipulation tasks. As shown in Figure~\ref{fig:demo-video-gen}, GE-Base generates multi-view videos that accurately reflect diverse language instructions. The results highlight the model’s ability to maintain spatial consistency across views, preserve background and scene structure, and produce stable, step-by-step execution aligned with the instruction semantics. Further analysis of video generation quality is provided in the benchmark section~(Section~\ref{sec:bench}).

\section{GE-Act: World Action Model}  
Bridging high-level world modeling and low-level control is essential for deploying vision-language foundation models in embodied robotics. We present \textbf{GE-Act}, a plug-and-play world action module that augments the fast LTX-Video–based GE-Base foundation model with a lightweight 160M-parameter autoregressive action decoder. GE-Act translates multimodal latent representations—conditioned on multi-view visual observations and language instructions—into temporally structured action policies, enabling instruction-following behaviors without explicit video generation. This architecture tightly couples perception and control, providing a scalable and efficient solution for real-time robotic manipulation across diverse environments.

\begin{figure}[t]
\begin{center}
\centerline{\includegraphics[width=1\linewidth]{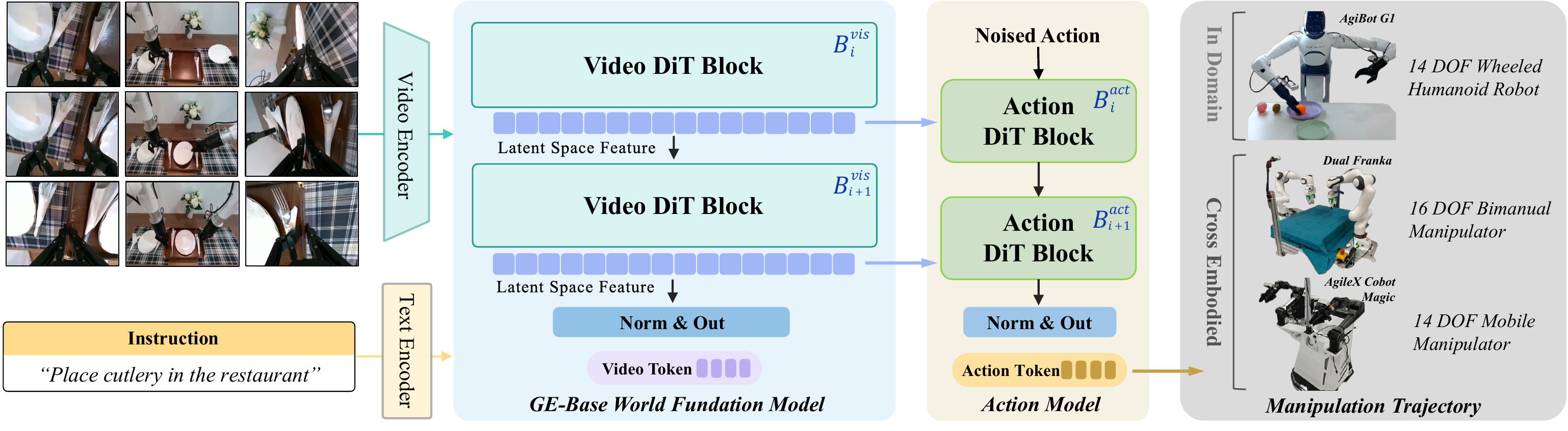}}
        \caption{\textbf{Overview of the GE-Act World Action Model.} GE-Act extends the GE-Base foundation model by incorporating a parallel action branch that converts visual latent representations into structured action policy trajectories. It follows the same block design and depth as GE-Base but reduces the hidden dimensions to improve efficiency. Visual latent features are integrated into the action pathway through a cross-attention mechanism, ensuring the semantic grounding of actions. Final action predictions are generated using a diffusion-based denoising flow-matching pipeline, refining noisy action predictions into coherent action trajectories.}
        \label{fig:framework-ge-act}
    \end{center}
\end{figure}

\subsection{Basic Architecture}

GE-Act is a plug-and-play world action module that extends the GE-Base foundation model to enable instruction-conditioned robotic control. Architecturally, it operates in parallel with the visual backbone of GE-Base, adopting an autoregressive DiT-based design that transforms latent visual representations into temporally structured action policies. This integration bridges high-level perceptual understanding with low-level motor execution, supporting seamless policy generation from multi-view visual observations and language instructions.

As shown in Figure~\ref{fig:framework-ge-act}, GE-Act preserves structural alignment with GE-Base by mirroring its DiT block depth while employing a reduced hidden dimension to ensure computational efficiency. At each step, the foundation model processes visual tokens derived from initial observations $\mathbf{x}_0$ and sparsely sampled historical frames $\hat{\mathbf{x}}_{t-1}$, conditioned on instruction embeddings $\mathcal{T}(q)$:

$$
\mathbf{v}_i = \mathcal{B}^{\text{vis}}_i(\mathbf{v}_\text{in}, \mathcal{T}(q)),
$$

where $\mathbf{v}_\text{in}$ denotes the input visual tokens, and $\mathcal{B}^{\text{vis}}_i$ represents the $i$-th visual DiT block in GE-Base.

Simultaneously, the action pathway in GE-Act processes noise-initialized action tokens $\mathbf{z}_\text{act}$ via a corresponding set of action-specific transformer blocks $\mathcal{B}^{\text{act}}_i$, incorporating relevant contextual information through cross-attention:

$$
\mathbf{a}_i = \mathcal{B}^{\text{act}}_i(\mathbf{z}_\text{act}, \text{CrossAttn}(\mathbf{z}_\text{act}, \mathbf{v}_i)),
$$

where $\mathbf{a}_i$ denotes the output action representation.

This modular architecture allows GE-Act to operate entirely within the latent feature space, enabling control inference without requiring explicit video generation during deployment. When integrated into real-world systems, the model can directly consume live perceptual inputs, maintaining policy consistency through a closed-loop formulation.

\subsection{Training Procedure}
We adopt a two-stage training paradigm inspired by standard vision-language-action (VLA) manipulation frameworks, consisting of task-agnostic pretraining followed by task-specific adaptation.

\begin{figure}[t]
    \begin{center}
\centerline{\includegraphics[width=1\linewidth]{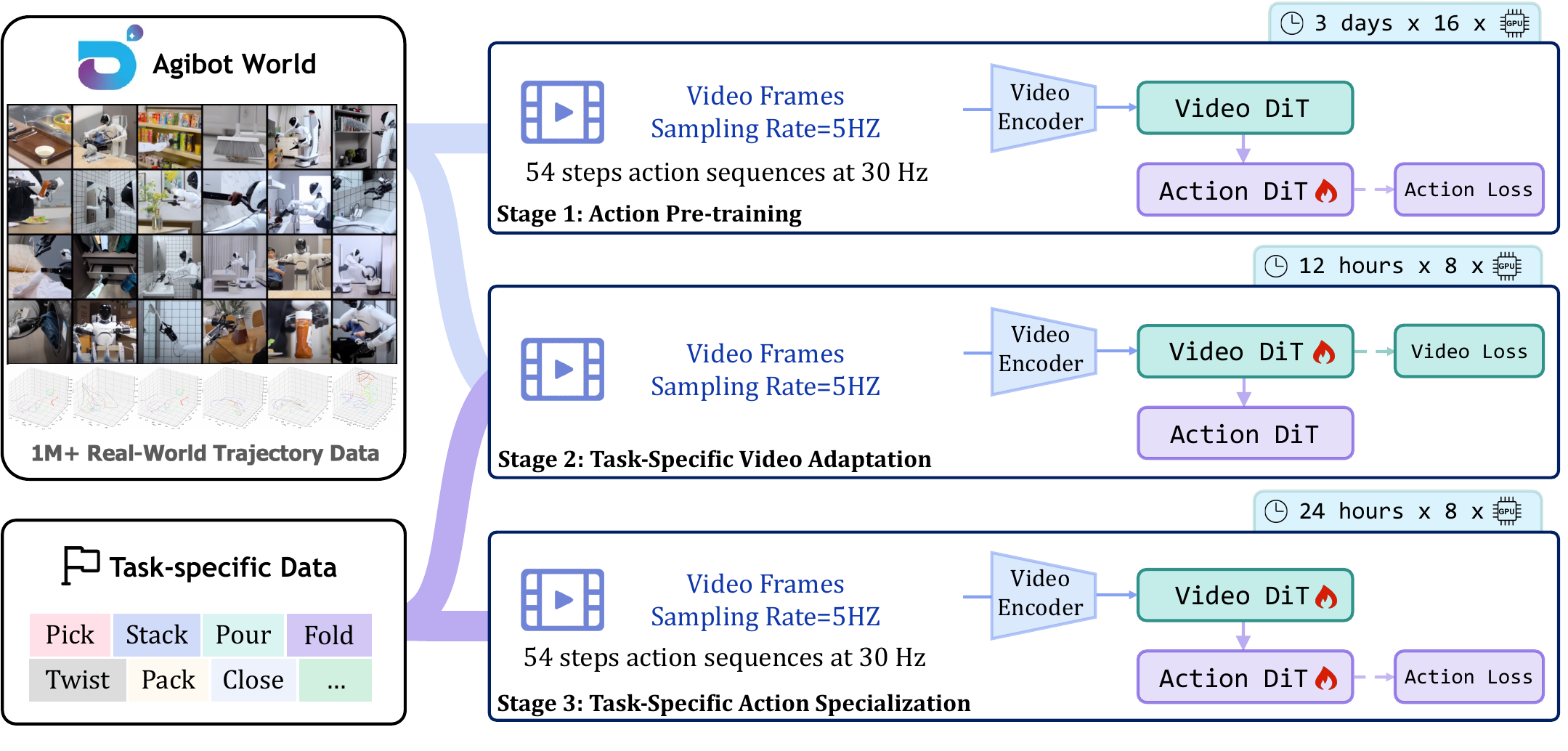}}
        \caption{\textbf{Overview of the GE-Act Training Pipeline.}
GE-Act is derived from the GE-Base foundation model through a three-stage training process utilizing text–video–policy triplets from the AgiBot-World-Beta dataset. The first stage performs action-space pretraining, where the visual backbone is optimized to project video sequences into a latent action policy space. Subsequently, a two-stage task adaptation procedure is conducted to specialize the model for diverse downstream tasks. In this phase, the video encoder is first adapted using task-specific visual data, followed by fine-tuning of the action head using corresponding control signals.
}
\vspace{-5mm}
        \label{fig:train-action}
    \end{center}
\end{figure}

\noindent\textbf{Pre-training.}
During the pretraining phase of the action model, we leverage the AgiBot-World-Beta dataset to specialize the pretrained visual-linguistic representations for action policy learning. The world model $\mathcal{W}$ is initialized with fixed parameters from GE-Base-LF to retain its spatiotemporal and semantic priors, while only the parameters of the action decoding module are updated. To reduce computational overhead, video generation is disabled during training. Instead, low-frame-rate visual memory sequences, consisting of four frames sampled at 5 Hz, are used as conditioning inputs, while the model predicts high-frequency action sequences comprising 54 steps at 30 Hz. The training is supervised solely by ground-truth action trajectories, enabling the model to learn control-relevant dynamics entirely within the pretrained latent space. This process completes in approximately three days on a cluster of sixteen NVIDIA A100 GPUs.

\noindent\textbf{Task-specific adaptation tuning.}
To adapt the pretrained model for downstream robotic tasks, we employ a two-stage fine-tuning pipeline comprising video adaptation and action specialization, aimed at aligning general visual-linguistic representations with task-specific execution requirements. During the video adaptation phase, we update only the video generation components of the world model $\mathcal{W}$, keeping the remaining parameters frozen. Fine-tuning is conducted on a composite dataset consisting of the full AgiBot-World corpus and a task-specific subset, with the latter upweighted by a factor of 10 to strengthen task alignment without sacrificing generalization. The sampling protocol is consistent with that used in GE-Base-LF to preserve temporal coherence. This phase is completed using 8 NVIDIA A100 GPUs over approximately 12 hours. In the subsequent action specialization phase, the full model—including both the GE-Base backbone and the action module—is fine-tuned exclusively on task-specific data to capture fine-grained control dynamics. This procedure mirrors the action pretraining setup and follows the same sampling strategy to ensure temporal and control-level consistency. This stage is trained using 8 NVIDIA A100 GPUs for approximately 36 hours.

\subsection{Asynchronous Inference}

To bridge the temporal gap between visual processing and motor control, we introduce the Slow-Fast Asynchronous Inference mode, which optimizes computational efficiency by exploiting asymmetries at two critical levels: denoising complexity and target frequency.

\noindent \textbf{Asymmetric Denoising Strategy.} Our inference pipeline allocates computational resources based on the distinct requirements of each modality. The video DiT performs a single flow-matching denoising step per inference pass to generate visual latent tokens, which are then cached and reused throughout the action generation phase. The action model, requiring higher temporal resolution for precise control, executes five denoising steps, all conditioned on the same cached visual representations. This approach ensures that the forward pass for 54 steps is completed in $200$ms on an onboard NVIDIA RTX 4090 GPU mounted on a real-world robot, ensuring real-time inference capability.

Beyond improving the denoising process, we exploit the inherent frequency mismatch between visual perception and motor control. The video DiT operates at 5 Hz, while the action model runs at 30 Hz, resulting in a temporal resolution ratio of 1 to 6. This decoupling enables sparse video prediction alongside dense action generation. By representing only selected future video frames, we significantly reduce the dimensionality of the video latent space, eliminating the need to process high-frequency visual sequences. This design allows the video DiT to operate efficiently in a compact latent space, while the action model retains the full temporal resolution necessary for accurate and responsive control.

This dual-level optimization delivers substantial advantages for both training and deployment. During training, we eliminate the typical bottleneck caused by video loading and decoding by initializing hidden states with random Gaussian noise, optimizing the training process for large-scale video models. In deployment, the combination of single-step video denoising and reduced latent dimensionality enables efficient real-time operation on robotic hardware, facilitating the seamless integration of video generation and action execution.

\subsection{Action Planning via GE-Act on AgiBot G1}

\begin{figure}[h]
    \begin{center}
\centerline{\includegraphics[width=1\linewidth]{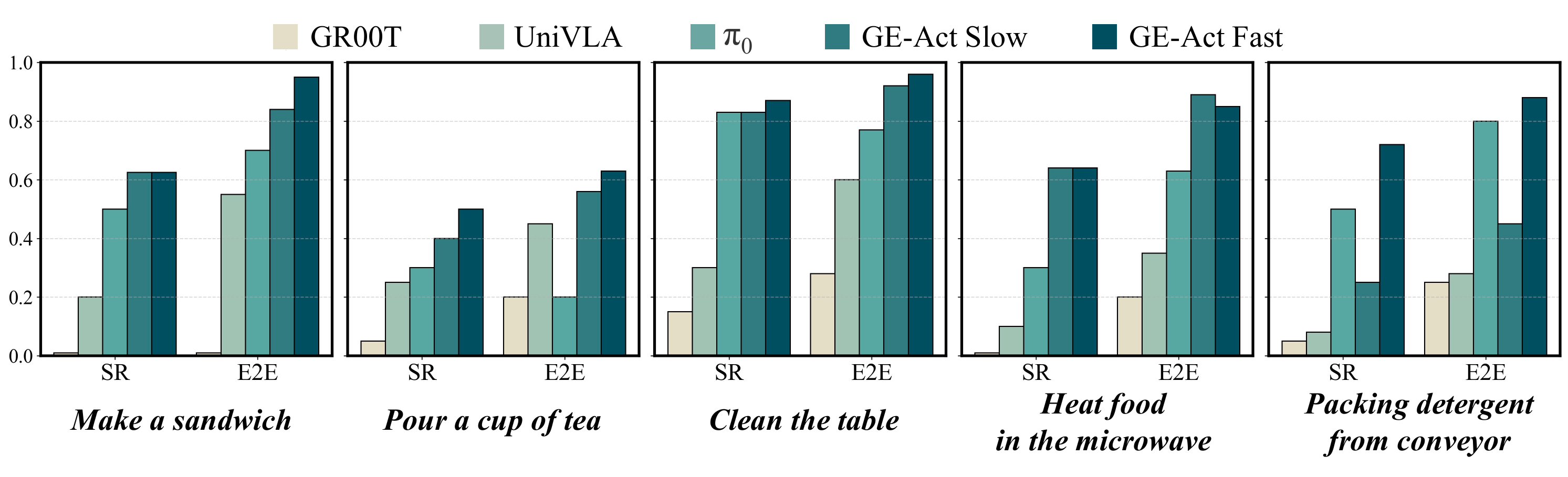}}
        \caption{\textbf{Comparison of Task-Specific Real-World Robotic Manipulation Performance on the AgiBot G1 Platform.}
        We compare GE-Act with state-of-the-art VLA baselines across multiple real-world dual-arm robotic tasks, using two evaluation metrics to assess performance.
}
\vspace{-5mm}
        \label{fig:exp-bar-sr-e2e}
    \end{center}
\end{figure}
To rigorously assess the effectiveness of our approach in real-world robotic manipulation, we conduct extensive evaluations across five representative tasks, each designed to test distinct aspects of control precision, task complexity, and generalization. These include: (1) \emph{Make a sandwich}: sequentially assembling bread, bacon, lettuce, and bread, which tests multi-object coordination, spatial reasoning, and procedural task execution; (2) \emph{Pour a cup of tea}: involving grasping, precise pouring, and repositioning a teapot, highlighting the need for fine-grained motion control and dexterity in fluid manipulation; (3) \emph{Clean the table}: requiring the robot to grasp a wiper and perform consistent wiping motions to remove surface stains, evaluating trajectory stability and compliant force application; (4) \emph{Heat food in the microwave}: operating a microwave door, inserting a bowl, and interacting with buttons, challenging the system’s ability to handle articulated objects and multi-stage interface operations; (5) \emph{Pack laundry detergent}: grasping moving detergent bags from a conveyor belt and placing them into boxes, designed to assess dynamic perception, motion tracking, and industrial-scale manipulation. These tasks span both household and industrial settings, providing a comprehensive benchmark for evaluating instruction-conditioned control, temporal grounding, and closed-loop execution capabilities.

\begin{figure}[t]
    \begin{center}
\centerline{\includegraphics[width=1\linewidth]{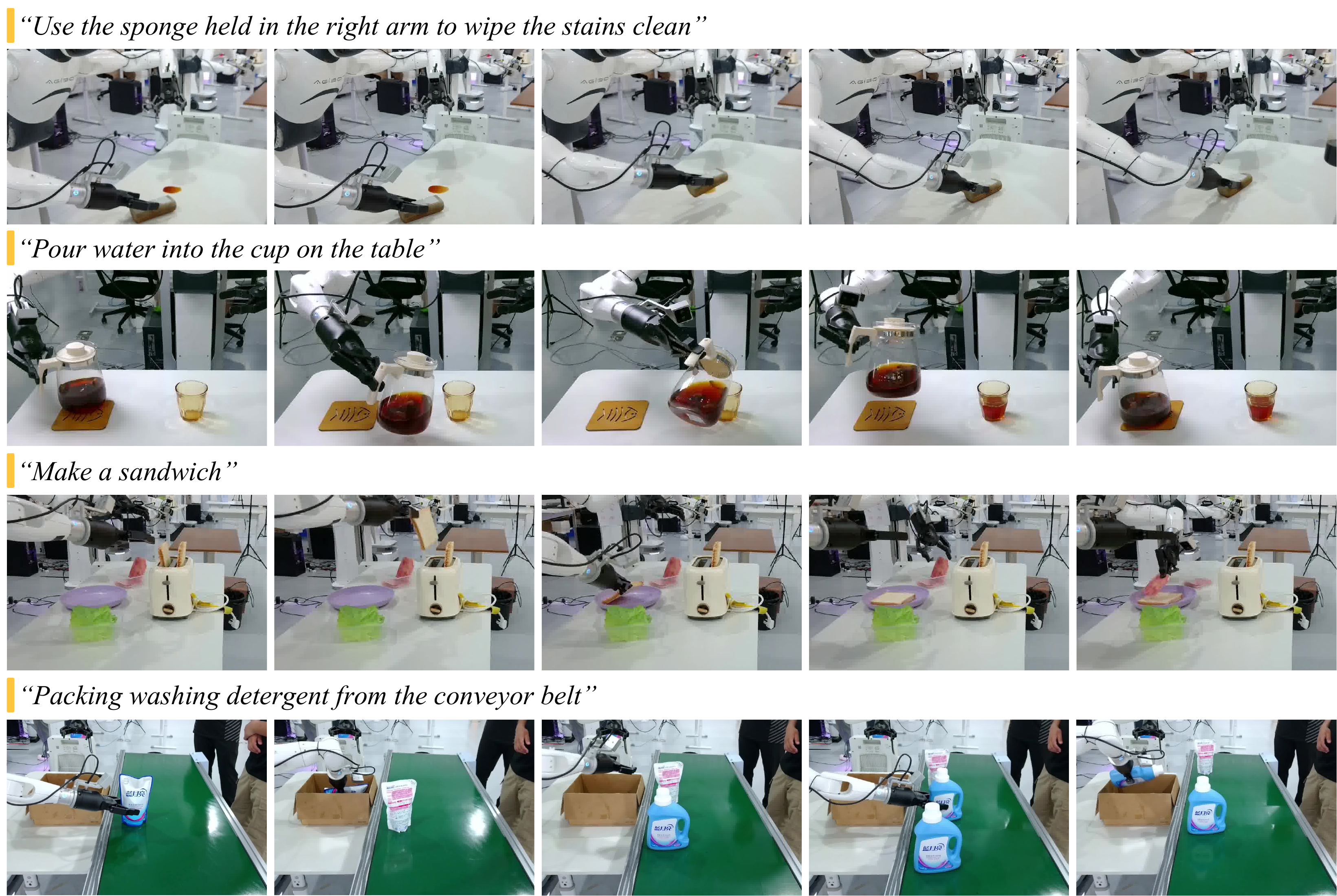}}
        \caption{\textbf{Visualization of Real-World Robotic Manipulation on AgiBot G1 via GE-Act.}
Conditioned on natural language instructions, GE-Act generates and executes action policies on the AgiBot G1 platform. The visual samples demonstrate the model's capability to produce consistent, reliable, and contextually appropriate manipulation behaviors, showcasing its robustness and effectiveness in real-world environments.}
\vspace{-3mm}
        \label{fig:demo-AgiBot G1}
    \end{center}
\end{figure}

 \noindent\textbf{Evaluation Protocols.} We employed two evaluation metrics to assess performance: Step-wise Success Rate (SR) and End-to-End Success Rate (E2E). The SR metric evaluates each sub-step independently and computes the overall success as the ratio of successfully completed sub-steps to total sub-steps, providing fine-grained insight into partial task completion. In contrast, the E2E metric evaluates only the final outcome of the complete task, allowing multiple attempts for individual sub-steps during execution, which better reflects real-world deployment scenarios where robots can recover from intermediate failures.

\noindent\textbf{Performance Comparison on the AgiBot G1 Platform.}
We benchmark GE-Act against two leading VLA-based robotic manipulation models: UniVLA~\citep{bu2025univla}, the state-of-the-art method on the LIBERO benchmark~\citep{liu2023libero}, and GR00T N1~\citep{bjorck2025gr00t}, a large-scale VLA foundation model. All models are evaluated on the AgiBot G1 platform following the same task protocols and using identical task-specific teleoperated demonstrations for fine-tuning. As shown in Figure~\ref{fig:exp-bar-sr-e2e}, GE-Act consistently outperforms baseline models across both SR and E2E metrics on a range of real-world daily manipulation tasks. This performance gain is attributed to the pretrained GE-Base world foundation model, which supplies strong spatiotemporal priors and precise visual-language grounding, enabling more efficient and robust adaptation to diverse downstream manipulation scenarios.

We further validate this design through two operational modes: the standard mode, which synchronizes visual and action updates, and the fast mode, which leverages temporal abstractions for improved efficiency. As demonstrated in Figure~\ref{fig:exp-bar-sr-e2e}, the fast mode achieves comparable or superior performance across various manipulation tasks, particularly excelling in latency-sensitive scenarios such as dynamic object tracking and reactive grasping. Notably, on short-horizon tasks like "Packing detergent from conveyor," which require rapid action generation, the fast model significantly outperforms the standard model.  Qualitative results are visualized in Figure~\ref{fig:demo-AgiBot G1}, showcasing GE-Act’s ability to execute complex manipulation tasks precisely and reliably in real-world settings, directly conditioned on natural language instructions.

\textbf{Analysis.} To systematically analyze our GE model, we conduct real-world robotic manipulation experiments on the AgiBot-G1 platform. We select a stable and controllable task, \emph{"grasping a red cylinder from the table and placing it into a paper cup with fixed positions"}, using a dataset of 305 demonstrations. All models are trained for 40,000 steps under the same protocol. Our analysis focuses on the role of pretraining in action policy prediction, comparing general video pretraining with in-domain embodied pretraining (AgiBot-World-Beta). As shown in Table~\ref{tab:arch_abla}, training from scratch or adapting from a general video model such as LTX-Video yields near-zero success. 
\begin{wrapfigure}{r}{0.50\textwidth}
\vspace{-5pt}
\begin{minipage}[h]{1\linewidth}
\centering
\captionof{table}{Analysis of Pre-training. ‘S’ denotes inclusion of robot state; ‘VidAW’ indicates initialization from GE-Base, ‘VidAda’ indicates task-specific video adaptation.} 
\vspace{-8pt}
\footnotesize
\renewcommand\tabcolsep{6pt}
\renewcommand\arraystretch{1.2}

\resizebox{1\linewidth}{!}{
\begin{tabular}{>{\centering\arraybackslash}p{1cm}>{\centering\arraybackslash}p{1.3cm}|cc|cc}
\rowcolor{cadetblue!20}
& & 
\multicolumn{2}{c|}{\textbf{E2E}} &
\multicolumn{2}{c}{\textbf{SR}} \\
\rowcolor{cadetblue!20}
\multirow{-2}{*}{\hspace{-0.1cm}\textbf{VidAW}} &
\multirow{-2}{*}{\hspace{-0.1cm}\textbf{VidAda}} &
 \textbf{w/ $\mathcal{S}$} & \textbf{w/o $\mathcal{S}$} & \textbf{w/ $\mathcal{S}$} & \textbf{w/o $\mathcal{S}$} \\
\hline
\textcolor{darksalmon}{\ding{55}} & \textcolor{darksalmon}{\ding{55}} & 0.15 & 0.30 & 0.05 & 0.11 \\
\rowcolor{gray!10}
\textcolor{darksalmon}{\ding{55}} & \textcolor{forestgreen}{\ding{51}} & 0 & 0.05 & 0 & 0 \\
\textcolor{forestgreen}{\ding{51}} & \textcolor{darksalmon}{\ding{55}} & 0.81 & 0.49 & 0.64 & 0.26 \\
\rowcolor{gray!10}
\textcolor{forestgreen}{\ding{51}} & \textcolor{forestgreen}{\ding{51}} & \textbf{0.89} & 0.37 & \textbf{0.76} & 0.37 \\
\end{tabular}
}
\label{tab:arch_abla} 
\end{minipage} 
\vspace{-5mm}
\end{wrapfigure} 
In contrast, in-domain pretraining achieves 64 SR and 81 E2E, which further improve to 76\% and 89\% when combined with general video pretraining. We further validate the effectiveness of incorporating robot state as input, which yields additional performance gains. 
However, when applied directly to general video-pretrained models, the inclusion of state information reduces performance due to short-cut learning effects. These results demonstrate that the GE-Base pretrained world model offers strong representations and serves as a solid foundation for action policy prediction.

\begin{figure}[t]
    \begin{center}
\centerline{\includegraphics[width=1\linewidth]{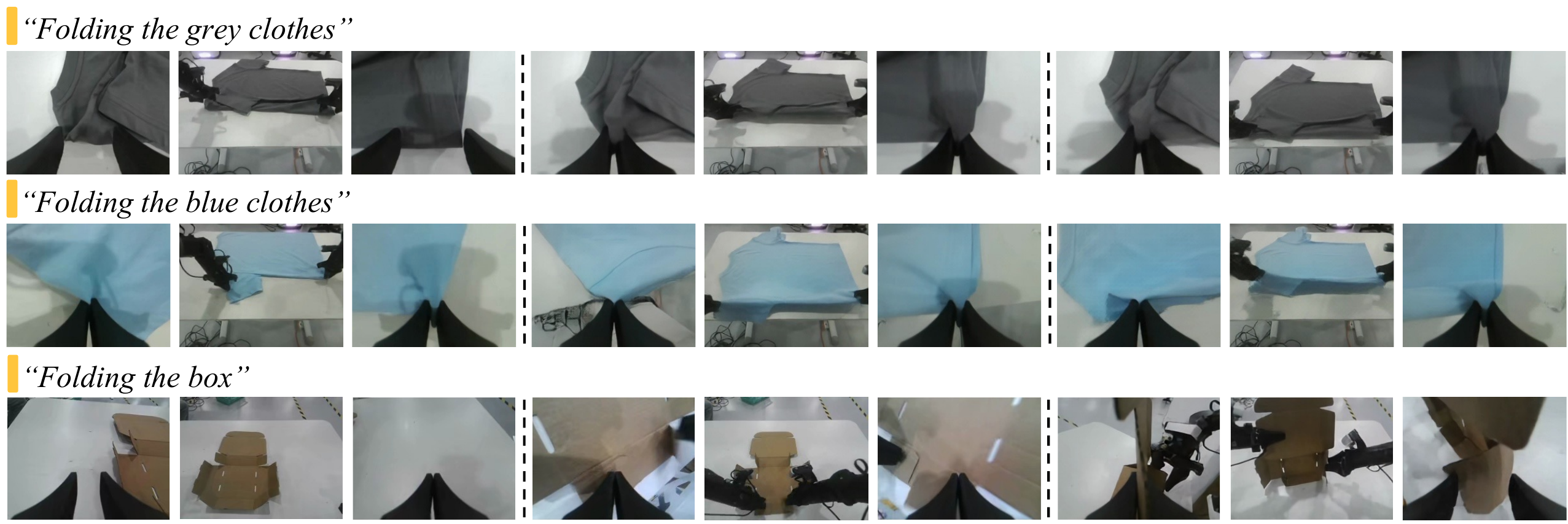}}
        \caption{\textbf{Multi-View Video Generation on the Agilex Cobot Magic Robotic Platform by GE-Base.} Visualization of instruction-conditioned video generated by GE-Base for two complex folding tasks on the cross-embodiment Agilex Cobot Magic robot. Each row displays temporally sampled frames from a multi-view sequence.}
        \vspace{-5mm}
        \label{fig:video-gen-cross-embody}
    \end{center}
\end{figure}

\section{Cross-Embodiment Generalization with Genie Envisioner}

Beyond validating GE on the in-domain AgiBot G1 platform, we assess GE’s ability to generalize across different embodiments, an essential step toward developing a versatile robotic foundation model. Specifically, we evaluate GE on two widely used robotic platforms in manipulation research: the Franka arm and the Agilex Cobot Magic system, as well as a dual-arm simulator, RoboTwin. To maintain consistency with our dual-arm framework, all platforms are configured accordingly.. To maintain consistency with our dual-arm framework, both platforms are configured accordingly.

Due to differences in embodiment design and action space, directly deploying the pretrained GE models on new platforms is not feasible. To overcome this, we use a few-shot adaptation protocol, collecting a small number of high-quality teleoperated demonstrations for each task. These demonstrations are used to fine-tune both the GE and GE-Act models, enabling effective transfer and alignment with the new platforms. In addition to the standard task set on AgiBot G1, we evaluate GE on complex deformable object manipulation tasks such as \emph{"cloth folding"} and \emph{"box folding"}, selected for their real-world relevance and physical challenges. This framework provides a thorough evaluation of GE's transferability, robustness, and control precision across different robotic platforms

\subsection{Few-shot Adaptation}
To enable few-shot adaptation on a novel robotic embodiment, we adopt a two-stage task-specific fine-tuning strategy for GE-Act, as illustrated in the final two rows of Figure~\ref{fig:train-action}:
\begin{itemize}
    \item In the first stage, we adapt the visual generative component to the new embodiment domain by fine-tuning the video DiT module using a small set of newly collected instruction-conditioned video demonstrations. During this process, the CLIP~\citep{radford2021learning} and video encoders are kept frozen to preserve pretrained semantic and perceptual priors. This step enables the model to synthesize realistic, embodiment-consistent manipulation videos aligned with the new platform’s visual characteristics.
    \item In the second stage, we train a new action DiT module from scratch using task-specific teleoperated trajectories. Owing to fundamental discrepancies in the embodiment structure and action space semantics, the pretrained action decoder is not reused. Instead, we retain the GE-Base visual backbone and learn a new action head tailored to the control dynamics and interface of the novel robotic platform.
\end{itemize}
This two-stage adaptation pipeline facilitates effective transfer of both perceptual and motor capabilities, enabling high-fidelity video generation and accurate, instruction-driven policy inference under minimal data supervision.

\begin{figure}[h]
    \begin{center}
\centerline{\includegraphics[width=1\linewidth]{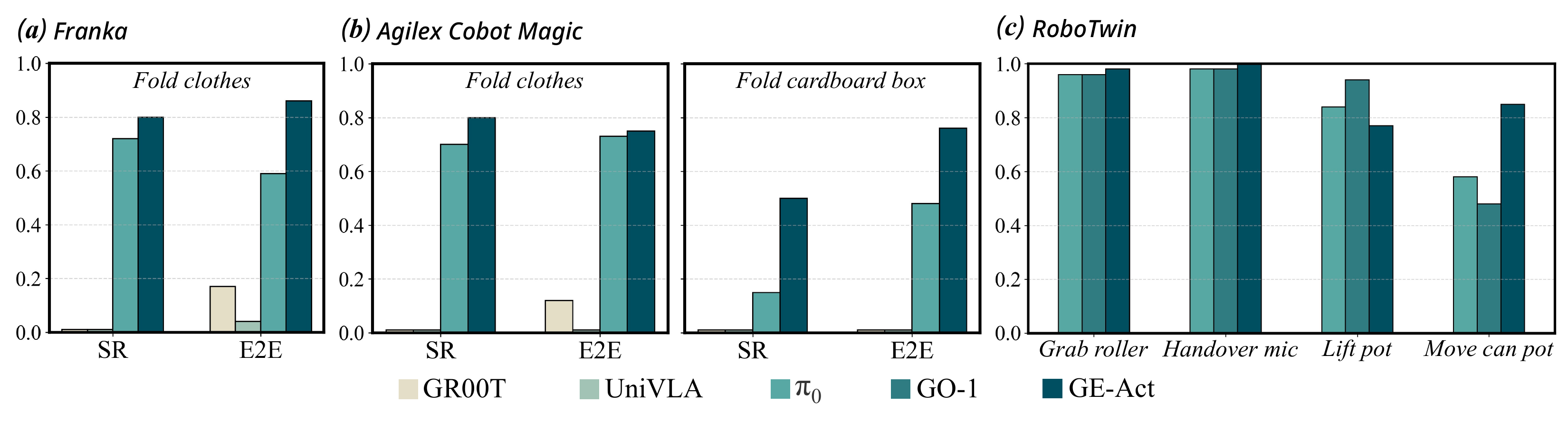}}
        \caption{Comparison of Task-Specific Manipulation Performance on Various Embodiments.}
        \label{fig:cross-embodie-compare}
    \end{center}
\end{figure}

\subsection{Generalization to Agilex Cobot Magic Embodiment}
We evaluate the generalization capability of GE on the Agilex Cobot Magic platform using two complex tasks: \emph{"box folding"} and \emph{"cloth folding"}. For each task, we collect 250 high-quality teleoperated demonstrations—approximately 1 hour of data—using the Aloha-based teleoperation system~\citep{fu2024mobile}. These demonstrations serve as the adaptation dataset to fine-tune both GE-Base and GE-Act.


As shown in Figure~\ref{fig:cross-embodie-compare}, we compare GE-Act with three state-of-the-art VLA models: GR00T N1~\citep{bjorck2025gr00t}, $\pi_{0}$~\citep{black2024pi_0}, and UniVLA~\citep{bu2025univla}. All models are fine-tuned on the same dataset for these tasks. The experimental results clearly show that GE-Act outperforms all three models. While UniVLA and GR00T N1 demonstrate strong capabilities in simpler tasks like pick-and-place, their lack of precision in positioning and task execution leads to failure when confronted with complex and fine-grained tasks, achieving a success rate of 0\%. Only with human intervention can UniVLA complete a few steps.  In contrast, $\pi_{0}$, known for its strong performance in deformable object manipulation, surpasses UniVLA and GR00T N1 in these areas. However, GE-Act significantly outperforms $\pi_{0}$ in complex, fine-grained deformable object manipulation tasks. This enhancement is primarily due to the GE-Base foundation model, which, through large-scale pretraining on real-world data, enables GE-Act to achieve better task adaptation and embodiment generalization. As a result, GE-Act delivers superior performance across a wide range of robotic platforms and manipulation scenarios.

\begin{figure}[t]
    \begin{center}
\centerline{\includegraphics[width=1\linewidth]{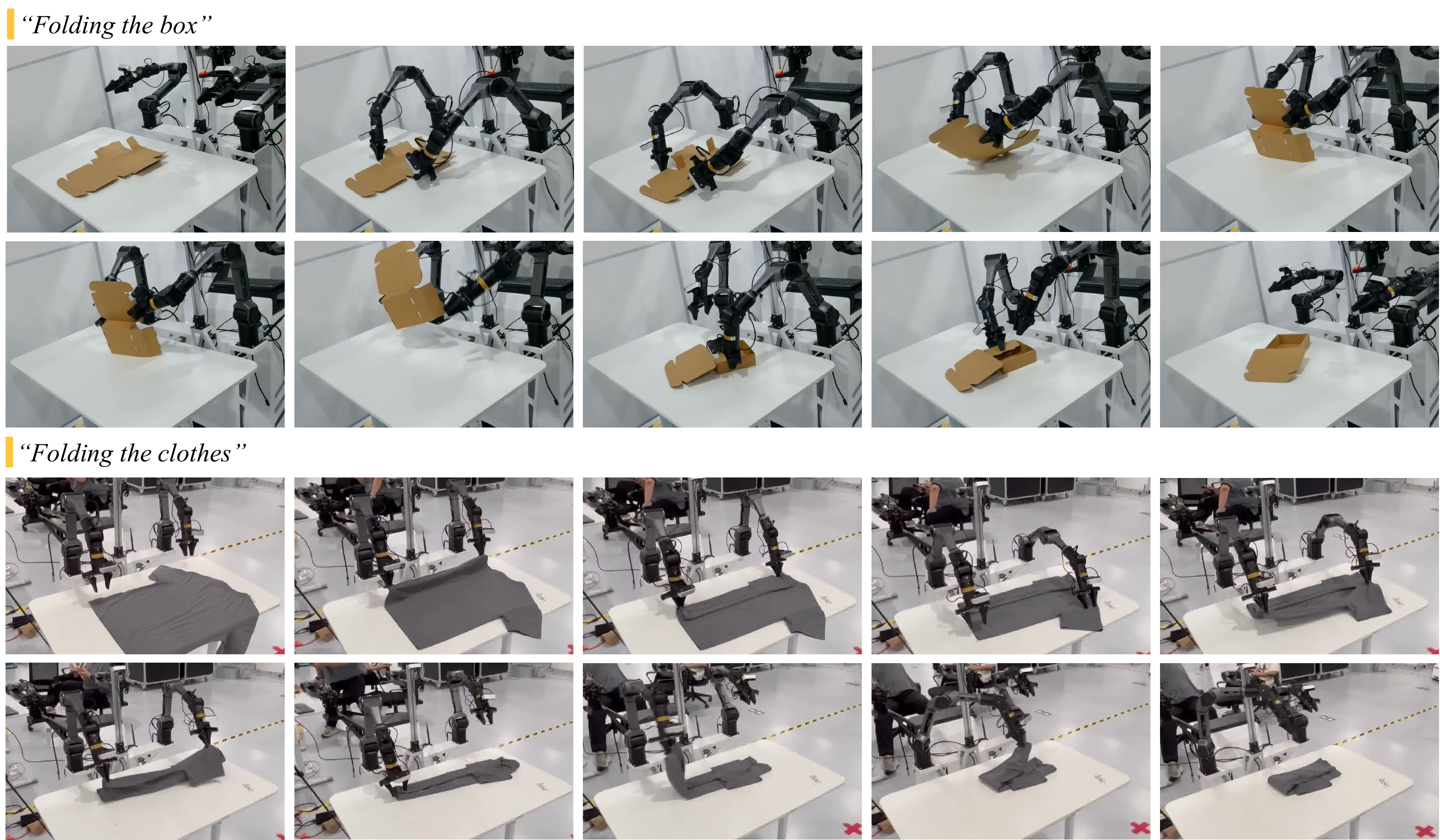}}
        \caption{\textbf{Visualization of Real-World Demonstrations with GE-Act on Agilex Cobot Magic Platform. }
This shows GE-Act adapted to a novel Agilex Cobot Magic embodiment, performing real-world robotic manipulation tasks, including cloth-folding and box-folding.}
        \label{fig:demo-cross-embody}
    \end{center}
\end{figure}

As illustrated in Figure~\ref{fig:video-gen-cross-embody}, our adapted GE-Act model generates coherent, instruction-conditioned multi-view videos for the cloth folding and box folding tasks. These videos accurately capture both rigid and non-rigid object dynamics with high fidelity. The results demonstrate strong consistency across different camera views and showcase GE-Act’s robust handling of complex object deformations. Furthermore, as shown in Figure~\ref{fig:demo-cross-embody}, we present real-world executions of the cloth folding and box folding tasks using the adapted GE-Act model. These results confirm GE-Act’s ability to complete the tasks with high precision and reliability on a novel robotic embodiment, further reinforcing the capacity of GE-Base to transfer effectively to new embodiments. This experiment solidifies the potential of GE-Base as a scalable, adaptable foundation for real-world embodied intelligence.

\begin{figure}[t]
    \begin{center}
\centerline{\includegraphics[width=1\linewidth]{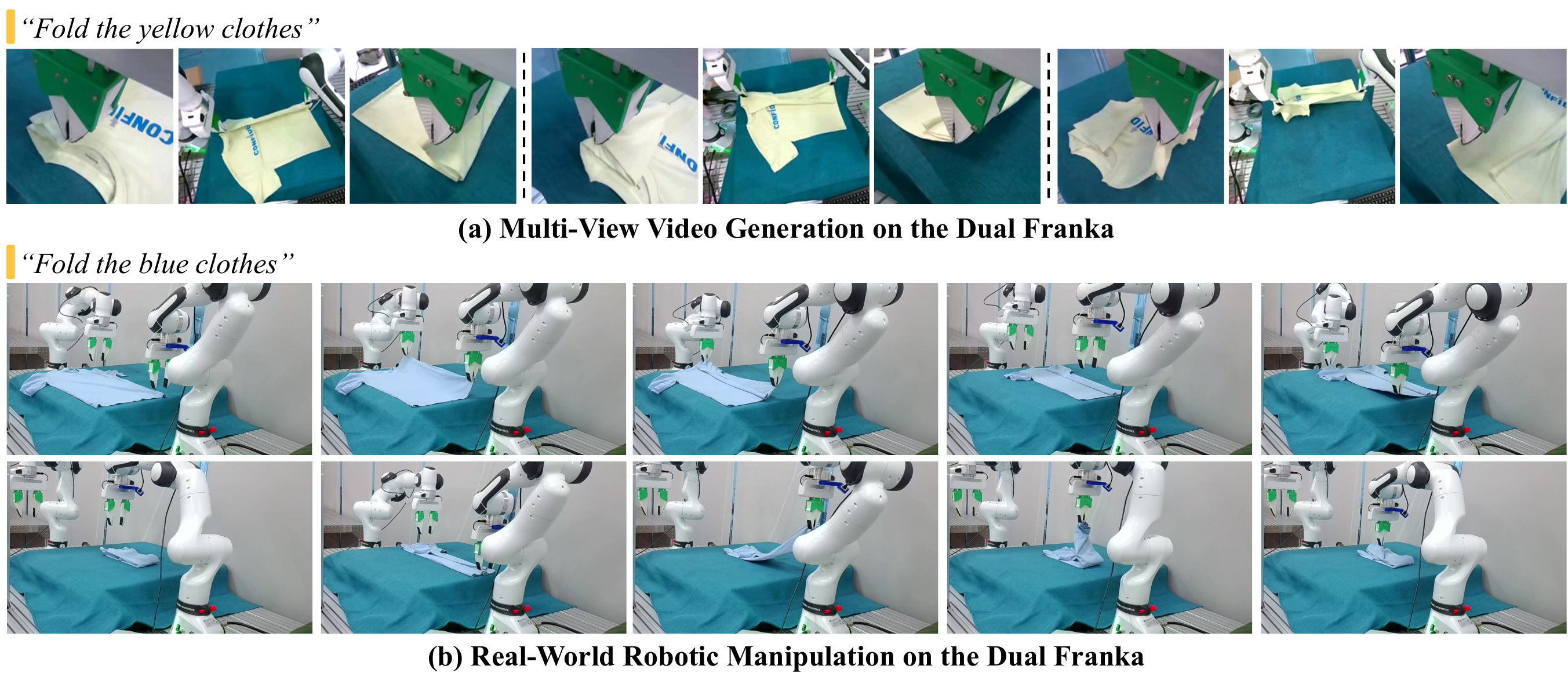}}
        \caption{\textbf{Visualization of Robotic Video Generation and Real-World Manipulation on Dual Franka via GE.} }
        \label{fig:demo-franka}
    \end{center}
\end{figure}

\subsection{Generalization to Dual Franka Embodiment}

We further evaluate GE’s cross-embodiment generalization on the Dual Franka platform by performing embodiment- and task-specific adaptation of GE-Act using 250 teleoperated episodes (approximately one hour) on the cloth folding task. Due to the absence of a dedicated teleoperation interface, data collection on Dual Franka is conducted using a simpler space-mouse-based control system. Consistent with the Agilex Cobot evaluation, we adopt GR00T N1~\citep{bjorck2025gr00t}, $\pi_{0}$~\citep{black2024pi_0}, and UniVLA~\citep{bu2025univla} as baselines, and fine-tune each on the 250-episode adaptation dataset. 
Figure~\ref{fig:demo-franka} illustrates the cloth folding task on the Dual Franka platform, including both the future-space video predictions by GE-Base and the real-world manipulation results executed by GE-Act. The results indicate that GE effectively models task-relevant visual dynamics and generalizes to new embodiments for precise manipulation. As shown in Figure~\ref{fig:cross-embodie-compare}, GE-Act consistently outperforms task-specific baselines in real-world execution on the Dual Franka platform, mirroring trends observed on the Agilex Cobot Magic. Notably, while $\pi_{0}$ and GR00T N1 were extensively trained on large-scale data from the Franka embodiment, GE-Act achieves superior performance with only one hour of adaptation data.

\subsection{Generalization to RoboTwin}


We further evaluate the cross-embodiment generalization on the dual-arm simulator RoboTwin~\citep{chen2025robotwin}. We adopt an all-in-one strategy, jointly fine-tuning GE-Act on four tasks using 200 demonstrations (50 per task), and directly evaluating this unified model across all tasks. In contrast, baseline methods~\citep{black2024pi_0,bu2025agibot} perform task-specific adaptation. As shown in Figure~\ref{fig:cross-embodie-compare}, GE-Act achieves better performance than $\pi_0$ and GO-1 on three of the four tasks, despite not using a one-task-one-model setting, and is only slightly behind VLA methods on lift pot. This minor gap may be attributed to task interference introduced by joint training.

\section{GE-Sim: World Simulator}
To support real-world-aligned evaluation and closed-loop control, we develop a video-based world neural simulator that generates temporally coherent visual predictions conditioned on robotic actions. This neural simulator enables embodied policy models to interact with a consistent visual environment, decoupled from physics-based constraints, and serves as a unified testbed for policy learning and generalization across diverse tasks. 

We realize this capability by extending the GE-Base foundation model into an action-conditioned simulator, GE-Sim. In this framework, action trajectories serve as the primary control signals driving video synthesis over time. To implement GE-Sim, we adopt two GE-Base architectures: the fast LTX-Video–based variant used in GE-Act, and the COSMOS2 2B–based variant for high-fidelity simulation and realistic video generation. To maintain visual consistency across generated frames, we incorporate a reference image encoded by a frozen CLIP image encoder as a lightweight style anchor. This reference is injected via cross-attention into each DiT block, complementing the spatial grounding provided by visual observations.

\begin{figure}[t]
    \begin{center}
\centerline{\includegraphics[width=1\linewidth]{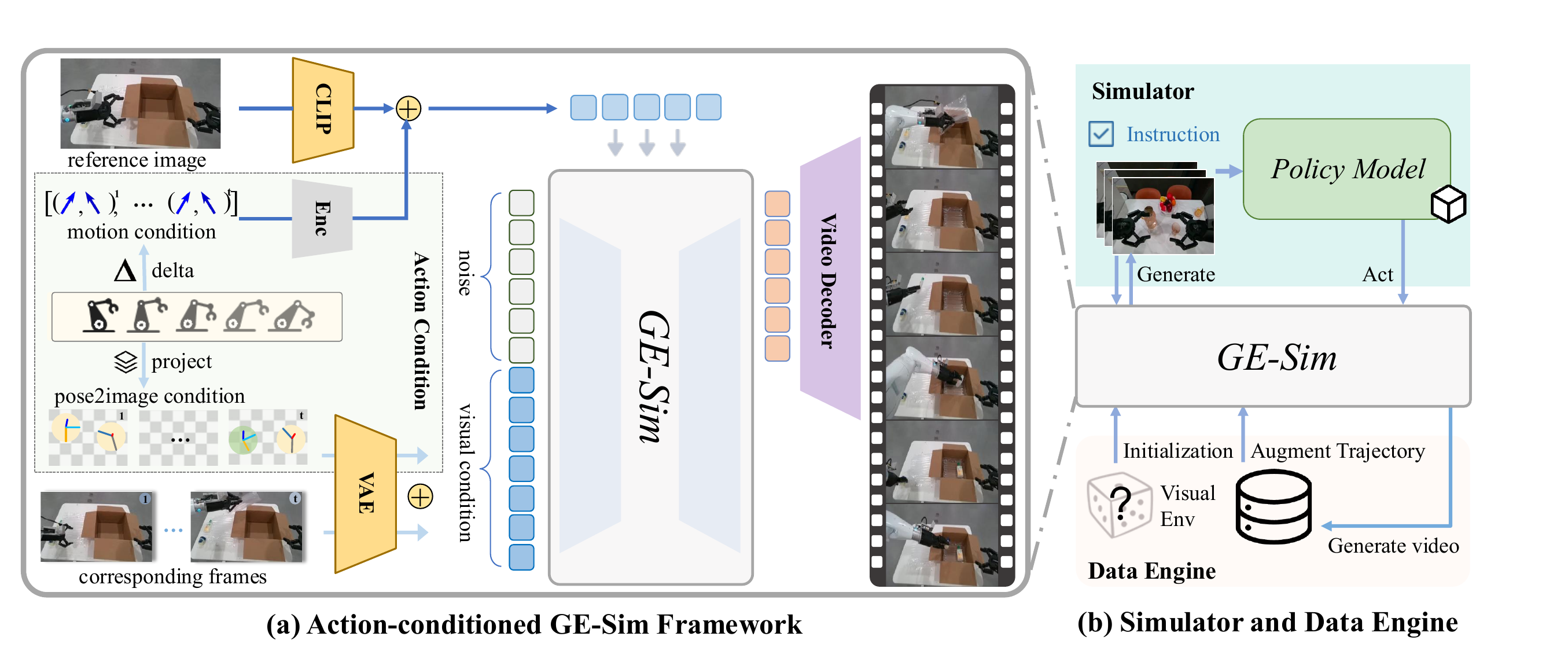}}
        \caption{\textbf{Overview of the GE-Sim World Simulator.}
(a) GE-Base is transferred into an action-conditioned video generator for simulating robotic behavior given predicted actions. Spatial pose conditions are projected into image space and fused with historical visual inputs, while temporal motion deltas are concatenated with a reference image to preserve style consistency and injected via cross-attention into the generation model.
(b) GE-Sim enables closed-loop policy evaluation and controllable data generation by producing action-conditioned video rollouts, supporting instruction-following and consistent trajectory replay under diverse visual contexts.}
        \label{fig:framework-enerverce-ac}
    \end{center}
\end{figure}

A fundamental challenge in this transformation is reconciling the semantic disparity between low-level control commands and the high-level latent representations encoded by the pretrained world model. To address this, as depicted in Figure~\ref{fig:framework-enerverce-ac}, we introduce a hierarchical action-conditioning mechanism that integrates structured action representations directly into the token space of GE-Base. This architecture preserves the model’s pretrained spatiotemporal semantics while enabling seamless interfacing with a wide range of policy models, thereby facilitating closed-loop, action-conditioned neural simulation with robust generalization to diverse robotic tasks.

\subsection{Hierarchical Action-conditioning Mechanism}

To ensure compatibility with diverse action policy models, we adopt a general representation of robotic trajectories. For a single manipulator, each control step is encoded as a 7-dimensional vector $[x, y, z, roll, pitch, yaw, o]$, where $(x, y, z)$ denotes the end-effector position, $(roll, pitch, yaw)$ its orientation (roll, pitch, yaw), and $o$ the gripper openness. In our dual-arm setup, the control signal per step is represented by a 14-dimensional vector formed by concatenating both arms' control vectors. Over a $K$-step horizon, the complete action trajectory is denoted as $\mathbf{A} \in \mathbb{R}^{K \times 14}$. To bridge this low-level control signal with the token-based input interface of the GE-Base foundation model, we propose a hierarchical action-conditioning mechanism that incorporates both spatial and temporal components.

\noindent\textbf{Pose2Image Conditioning.}
At each timestep $i$, the pose vector $a_i = [x_i, y_i, z_i, r_i, p_i, y_i, o_i]$ encodes the spatial position, orientation, and gripper state. The position $(x_i, y_i, z_i)$ is projected into pixel coordinates using calibrated camera intrinsics and extrinsics. The orientation $(r_i, p_i, y_i)$ is converted into a rotation matrix, whose orthonormal axes are also projected into the image plane to indicate directionality. The gripper openness $o_i$ is rendered on a unit circle, with shading intensity reflecting its state—lighter for open, darker for closed. Distinct color encodings differentiate the left and right arms. This process yields pose images $\mathbf{P}_i$ that are spatially aligned with the visual scene.

Each $\mathbf{P}_i$ is paired with its corresponding sampled history frame $\mathbf{I}_i$. Both are encoded using a shared video encoder $\mathcal{E}$, and their latent features are fused by element-wise addition:
\begin{equation}
\mathbf{v}_i = \mathcal{E}(\mathbf{I}_i) + \mathcal{E}(\mathbf{P}_i).
\end{equation}
The resulting fused token $\mathbf{v}_i$ captures both contextual visual semantics and explicit pose information, and is inserted into the visual token stream for downstream processing.

\noindent\textbf{Motion Vector Conditioning.}
To capture temporal dynamics, we compute motion deltas between consecutive end-effector poses. Let $\mathbf{a}_i = [\mathbf{p}_i, \mathbf{r}_i]$ denote the 6-DoF pose at timestep $i$, with $\mathbf{p}_i \in \mathbb{R}^3$ as position and $\mathbf{r}_i \in \mathbb{R}^3$ as orientation. The delta is given by:
\begin{equation}
\Delta \mathbf{a}_i = \mathbf{a}_i - \mathbf{a}_{i-1} = [\Delta \mathbf{p}_i, \Delta \mathbf{r}_i],
\end{equation}
which encodes both positional and orientational change. These deltas are encoded into motion tokens via a learnable encoder, concatenated with the reference image style token, and injected via cross-attention into each DiT block. This temporally-aware representation provides coherent motion priors to guide action-conditioned video generation within GE-Sim.

\begin{figure}[t]
    \begin{center}
\centerline{\includegraphics[width=1\linewidth]{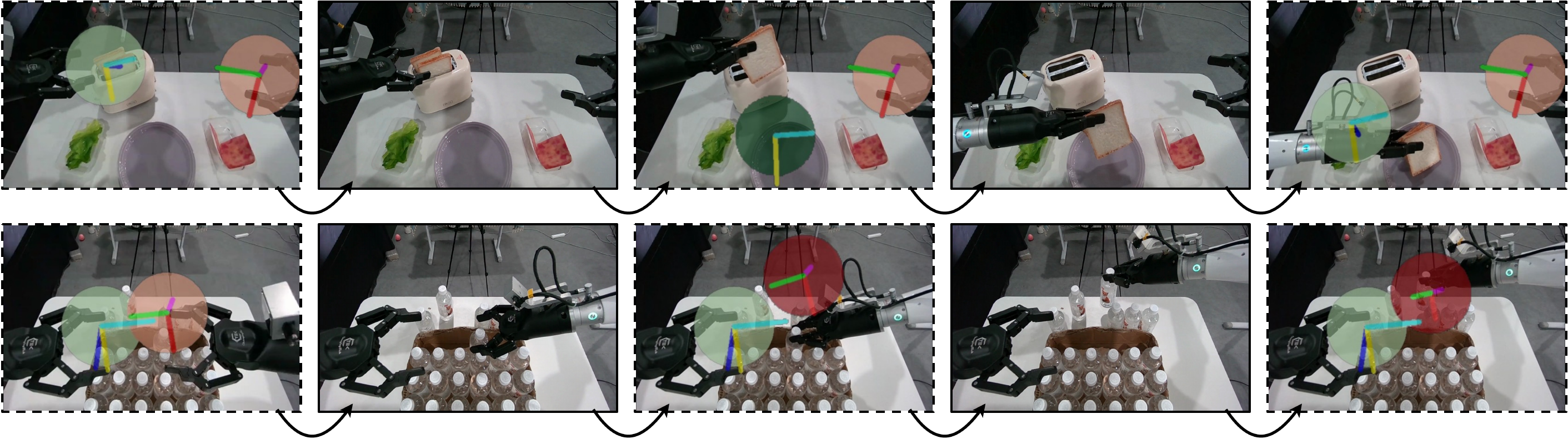}}
        \caption{\textbf{Visualization of Action-Conditioned Video Generation by GE-Sim.} Given a ground-truth action policy, we generate the corresponding next-frame prediction using GE-Sim. For each sample, we visualize the head-view image by overlaying the projected action target onto the current frame, alongside the predicted next frame, to illustrate the model’s spatial alignment with the intended control signal.}
        \label{fig:demo-video-gen-ac}
    \end{center}
\end{figure}

\begin{figure}[t]
    \begin{center}
\centerline{\includegraphics[width=1\linewidth]{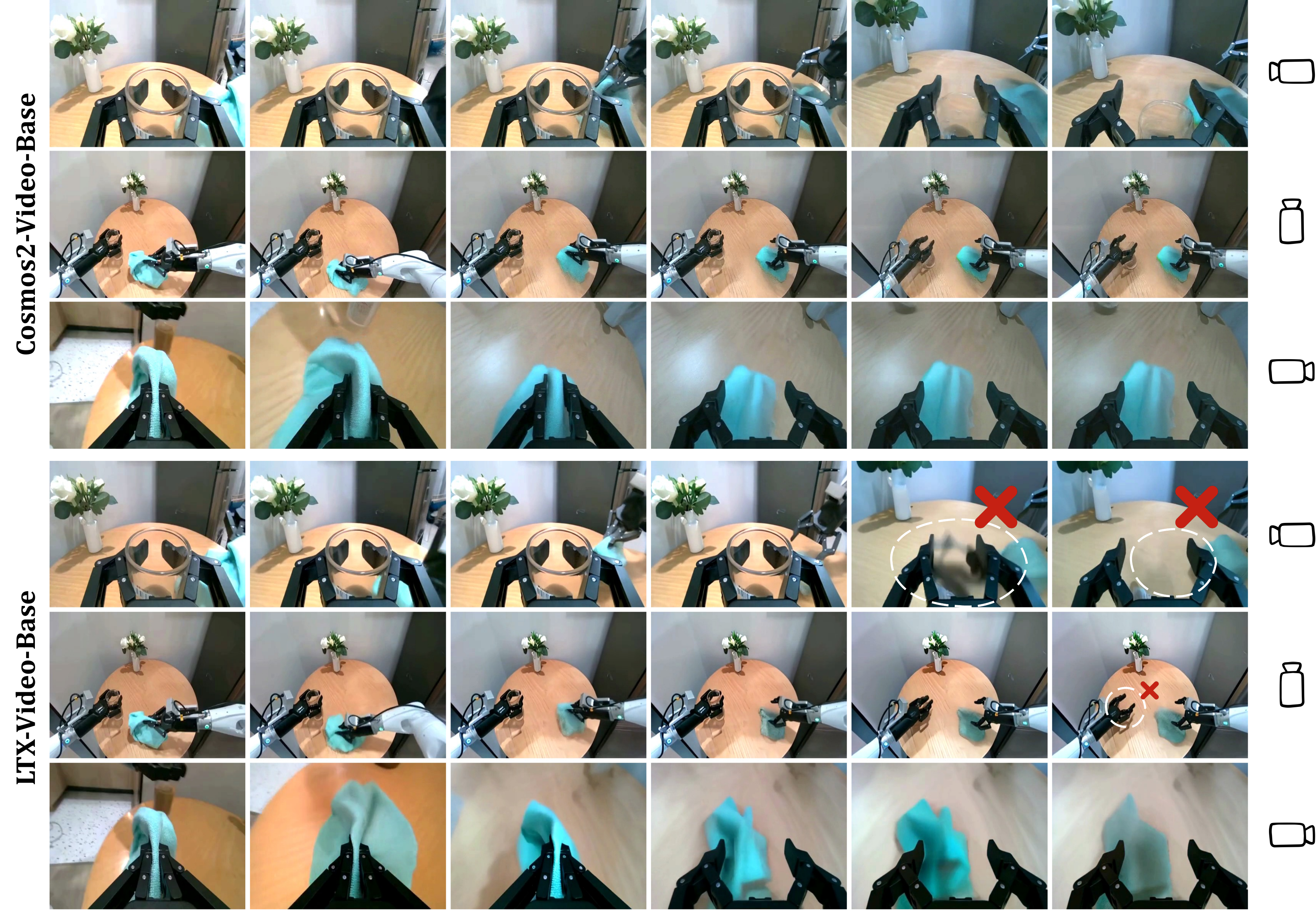}}
       \caption{\textbf{Comparison of Multi-View Action-Conditioned Generation Results between Two GE-Sim Variants.} 
Visualization of GE-Sim outputs based on different base models: COSMOS2 and LTX-Video, under identical action conditions.}
        \label{fig:demo-ge-sim}
    \end{center}
\end{figure}
\subsection{Training Procedure}

To ensure high-fidelity video simulation required for action-conditioned generation, GE-Sim is initialized from the high-temporal-resolution pretrained model GE-Base-MR, which offers fine-grained modeling of robotic dynamics. The model is subsequently trained on the full AgiBot-World-Beta dataset, using ground-truth action trajectories as conditioning inputs for video generation. To improve generalization and robustness, the training corpus is augmented with a diverse set of failure cases—including erroneous executions, incomplete behaviors, and suboptimal control trajectories—collected from both human teleoperation and real-world robotic deployments. During this phase, the VAE and CLIP encoders are kept frozen to preserve pretrained semantic and spatial priors, while the remaining parameters are optimized via a flow-matching loss applied over the predicted video representations.

\subsection{Action-conditioned Video Generation}
To evaluate the precision of action-conditioned video generation, we visualize simulation outputs from GE-Sim based on ground-truth control sequences. As shown in Figure~\ref{fig:demo-video-gen-ac}, each example presents the current observation frame overlaid with the projected target position of the next action, along with the corresponding predicted frame synthesized by the simulator. Across tasks and viewpoints, the generated end-effector motion consistently aligns with the spatial intent of the action input, demonstrating that GE-Sim can accurately translate low-level control commands into coherent visual predictions. Furthermore, we compare GE-Sim built on two base architectures under identical action-conditioning in Figure~\ref{fig:demo-ge-sim}. The COSMOS2-based variant exhibits higher visual fidelity and stronger temporal consistency than the LTX-Video–based model, confirming its superior capability in generating high-quality, action-aligned robotic simulations.

\subsection{Closed-Loop Simulation}
To support closed-loop evaluation of arbitrary policy models, GE-Sim functions as a video-based world simulator. Given a language instruction and initial visual observations, the policy model first takes these as input and outputs an action trajectory. GE-Sim then conditions on both the initial observations and the predicted action policy to generate a video chunk simulating the outcome of the action. This generated video is fed back into the policy model, along with the original instruction, to produce the next action step. This iterative process continues until the instruction is completed, enabling closed-loop simulation and real-world-aligned evaluation of policy models in a consistent and controllable visual environment.

Beyond policy evaluation, GE-Sim also functions as a versatile data engine. By executing the same action trajectory under different initial visual environments, it can generate diverse manipulation sequences reflective of varied contexts.

This video world simulator, grounded in real-world data, offers a compelling alternative to traditional physics simulators, achieving high visual fidelity while significantly reducing deployment costs. Crucially, it enables scalable, flexible simulation without requiring manual environment modeling. As such, GE-Sim lays the foundation for a new class of general-purpose, realistic, and low-cost world models that bridge learning and evaluation in embodied intelligence.

\section{EWMBench: Embodied World Model Benchmark}
\label{sec:bench}
An effective evaluation framework functions as a navigational instrument for scientific progress—establishing standardized criteria and fostering meaningful comparisons across methodologies. In the context of robotic world modeling, the ability to systematically assess whether a model faithfully captures the structure, dynamics, and semantics of embodied environments is essential for advancing the field. To this end, we propose the embodied world model benchmark, EWMBench, a comprehensive evaluation suite designed to measure both representational fidelity and practical utility of video-based world models in real-world robotic manipulation.

Beyond conventional benchmarks for general video generation that focus on visual fidelity, language alignment, or human preference, robotic manipulation videos introduce stricter structural constraints. In this domain, background layouts, object configurations, and embodiment structures (\emph{e.g.}, robot morphology) should remain invariant, while only the robot’s pose and interactions evolve in accordance with the instruction. EWMBench is designed with these domain-specific properties in mind, providing task-oriented metrics that assess visual-scene consistency, motion correctness, and semantic alignment and diversity, enabling more faithful and practical evaluation of world models in manipulation-centric scenarios. To support this, EWMBench comprises a high-quality real-world benchmark dataset and a suite of open-source evaluation tools, establishing a standardized framework for rigorously assessing the capabilities of video-based world models in manipulation-centric tasks.

\subsection{Benchmark Dataset}
The benchmark dataset is curated from the AgiBot-World-Beta test set by selecting 10 representative tasks spanning household and industrial domains. These tasks are characterized by well-defined operational goals and strong sequential dependencies, requiring procedural reasoning over affordances and action ordering. To ensure a fair evaluation, all selected tasks are disjoint from those used during the 1M-scale pre-training phase. Each task is decomposed into 4–10 atomic sub-actions, with every sub-action annotated with a step-level caption, enabling fine-grained alignment between video segments, action labels, and linguistic descriptions. For each task, we uniformly sample 100 video instances to construct a balanced and comprehensive evaluation set.

To promote diversity within each task, we implement a trajectory selection strategy based on spatial variation. Specifically, dual-arm end-effector trajectories are extracted and voxelized into 3D grids. A pairwise similarity matrix is computed using 3D Intersection-over-Union (IoU), and a greedy algorithm is employed to iteratively select the least-overlapping trajectories. This approach ensures broad coverage of motion patterns and minimizes redundancy within each task’s evaluation set.

\subsection{Evaluation Metrics}
We establish a unified evaluation framework to assess how accurately video-based world models capture the spatial, temporal, and semantic dynamics of robotic manipulation.

\noindent\textbf{Scene Consistency.} To evaluate the structural and visual coherence of generated videos, we introduce a scene consistency metric that assesses the stability of visual appearance, environment layout, and viewpoint alignment across time. Specifically, we propose a patch-level feature similarity metric computed over consecutive and initial frames. We first fine-tune a strong visual encoder, DINOv2~\citep{oquab2023dinov2}, on a robotic manipulation dataset to align its representation space with the embodied domain. For each frame, we extract patch-wise embeddings using this encoder. Then, cosine similarity is computed across corresponding patches between frames to quantify temporal consistency. Higher similarity scores reflect greater preservation of scene structure and camera viewpoint throughout the video sequence, indicating stronger spatial-temporal fidelity.

\noindent\textbf{Action Trajectories Quality.} To evaluate the quality of action trajectories executed in response to instructions, we manually annotate a reference trajectory for each instruction as the GT. For each generated video, a trained EEF detector is used to localize the gripper across frames and reconstruct the trajectory. Three video samples are generated per instruction, and the corresponding trajectories are extracted. Spatial alignment~(SA) is assessed using the Symmetric Hausdorff~(symH) Distance, which measures the maximum point-wise deviation between the generated trajectory $P$ and the ground truth $G$. To ensure higher scores indicate better alignment, we report the inverse of this value:
$$
\text{SA}_{\text{score}} = \frac{1}{d_{\text{symH}}(G, P) + \epsilon}.
$$
To account for generation variance, the trajectory with the lowest symH is selected for further evaluation. 

Temporal alignment~(TA) is then evaluated using Normalized Dynamic Time Warping (NDTW)~\citep{ilharco2019general}, which captures consistency in both sequence and timing between the generated and ground truth trajectories. To produce a positively correlated metric, we report the inverse of the NDTW distance:
    $$
\text{TA}_{\text{score}} = \frac{1}{d_{\text{NDTW}}(G, P) + \epsilon}
$$
Additionally, we introduce a Dynamic Consistency (DYN) metric to assess the realism of motion dynamics by comparing velocity and acceleration profiles between predicted and ground-truth trajectories. Specifically, we compute the Wasserstein distance $W(\cdot)$ between the respective time series, capturing distributional alignment without requiring strict temporal correspondence. To account for variations in motion amplitude and prevent instability in low-dynamic cases, we normalize each component using amplitude-aware ratios. The final score is defined as:

$$
\text{DYN}_{\text{score}} = \alpha \cdot \frac{\min(\Delta v^\text{gt}, \Delta v^\text{pred}) + \epsilon}{\max(\Delta v^\text{gt}, \Delta v^\text{pred}) + \epsilon} \cdot \frac{1}{W(v)} + 
\beta \cdot \frac{\min(\Delta a^\text{gt}, \Delta a^\text{pred}) + \epsilon}{\max(\Delta a^\text{gt}, \Delta a^\text{pred}) + \epsilon} \cdot \frac{1}{W(a)}
$$

where $\Delta v = \max(v) - \min(v)$, $\Delta a = \max(a) - \min(a)$, $\epsilon = 10^{-8}$, and $\alpha = 0.007$, $\beta = 0.003$. This formulation ensures that the metric reflects both dynamic fidelity and amplitude robustness.
This multi-level evaluation provides a comprehensive measure of spatial, temporal, and dynamic fidelity.

\begin{figure}[t]
    \begin{center}
\centerline{\includegraphics[width=1\linewidth]{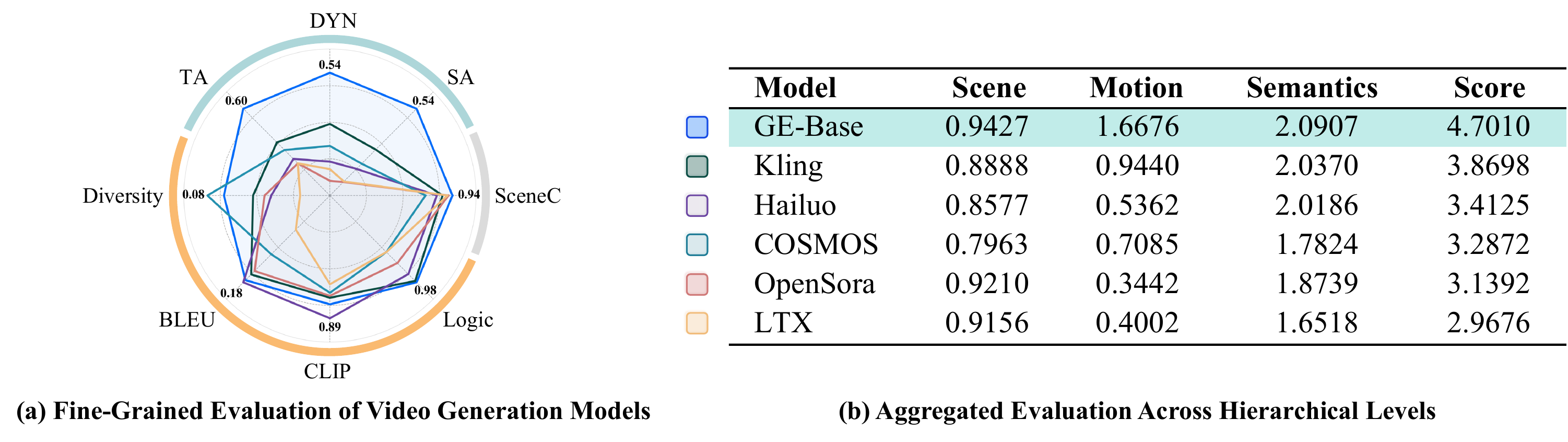}}
        \caption{\textbf{Comprehensive Evaluation of Video World Models for Robotic Manipulation.} Leveraging our EWMBench , we systematically evaluate a suite of video world models sourced from state-of-the-art general video generation and embodied world modeling approaches. All models are assessed under a unified text-and-image to video generation paradigm. Evaluation spans multiple levels, scene, motion and semantics, capturing visual fidelity, temporal coherence, and semantic grounding in diverse real-world robotic manipulation tasks.
}
        \label{fig:exp-8metric}
    \end{center}
\end{figure}

\noindent\textbf{Motion Semantics Metrics.} 
We evaluate motion semantics from two perspectives: semantic consistency and behavioral diversity. Semantic consistency assesses whether the generated manipulation behavior aligns with the intended task instruction, while diversity measures the model's ability to produce varied yet valid trajectories.  For semantic consistency, we adopt a multi-granularity evaluation framework based on the VLM, Qwen2.5-VL-7B-Instruct~\citep{bai2025qwen2}:
\begin{itemize}[leftmargin=*]
    \item \emph{Global-level alignment:} A VLM generates a compact summary caption for each generated video, which is then compared to the original task-goal instruction using BLEU scores to assess overall alignment between the video and intended task semantics.
    \item \emph{Key-step consistency:} To assess whether essential sub-tasks are correctly executed, the VLM generates step-by-step descriptions for both the generated and ground-truth manipulation videos. Consistency is measured by computing CLIP-based similarity between the corresponding steps in the two descriptions.
    \item \emph{Logical correctness:}  To identify violations of physical or commonsense constraints, we first prompt GPT to define a taxonomy of typical logical errors in robotic manipulation videos, such as hallucinated actions, object disappearances, or physically implausible motions. Then, a video-based VLM is used to detect the presence of these predefined error types in generated videos. Detected violations are explicitly penalized, encouraging the model to produce semantically accurate and physically coherent manipulation behaviors.
\end{itemize}

To assess the model’s capacity for generating varied outputs, we measure semantic diversity using CLIP-based global video embeddings. Specifically, we compute pairwise CLIP similarities between generated videos conditioned on the same instruction, and define the diversity score as  1 - CLIP similarity. Higher scores indicate greater semantic variability, reflecting the model’s ability to generalize beyond deterministic execution.

\subsection{World Model Evaluation}
To thoroughly evaluate the effectiveness of video-based world models for robotic manipulation, we establish a comprehensive evaluation framework, referred to as the "evaluation colosseum," enabling direct, comparative analysis across various model architectures. In this framework, we benchmark seven state-of-the-art video generation models, including Open-Sora~\citep{opensora}, Kling~\citep{Kuaishou2025Kling}, Hailuo~\citep{hailuo}, LTX-Video~\citep{HaCohen2024LTXVideo} and the scene-centric COSMOS~\citep{agarwal2025cosmos}. All models are evaluated under a standardized text-and-image-to-video generation paradigm, where natural language instructions and head-view visual observations condition video synthesis. Notably, the GE-Base model is built on the LTX-Video architecture, which enables it to focus on domain-specific tasks and leverage fine-tuned control.

As shown in Figure~\ref{fig:exp-8metric}, \textbf{GE-Base} consistently outperforms the baselines across multiple evaluation dimensions, with notable strengths in temporal alignment and dynamic consistency, two core metrics for generating action-plausible and temporally stable robotic behaviors. While performance in motion semantics is comparable to generic video generation models, GE-Base demonstrates much stronger control-aware generation fidelity, offering more precise and reliable task execution. This advantage is attributed to GE-Base's pretraining on large-scale robotic manipulation data, which better equips it to capture task-relevant spatial-temporal dynamics.

In comparison, Kling~\citep{Kuaishou2025Kling} achieves strong overall performance, particularly in robustness across general video generation tasks, but lacks the specialized understanding required for fine-grained control, which limits its performance on more complex robotic manipulation tasks. Hailuo~\citep{hailuo}, though proficient in zero-shot embodied scenarios, often generates cartoon-like outputs that compromise visual realism, limiting its applicability for real-world robotic manipulation. COSMOS~\citep{agarwal2025cosmos} and LTX-Video~\citep{HaCohen2024LTXVideo} models, while effective in human hand-centric tasks, struggle with adapting their semantic understanding to robotic contexts, and often produce inconsistent task execution. Notably, LTX-Video experiences abrupt scene transitions and a tendency to generate stationary states during action sequences, while COSMOS struggles with maintaining consistent viewpoints and camera control. Lastly, OpenSora~\citep{opensora} displays partial understanding of task scenes and action semantics, but frequently suffers from jittery robotic arm movements and generates static videos, particularly in more complex tasks.

These results highlight GE-Base's advantages in bridging high-level semantic understanding with low-level control execution. Its superior performance in temporal alignment, dynamic consistency, and task adaptation positions GE-Base as a leading model for real-world robotic manipulation.

\subsection{Simulation Evaluation}  
\begin{figure}[h!]
\centering
\begin{minipage}[b]{0.56\textwidth}
    \centering
    {
    \includegraphics[width=1\linewidth]{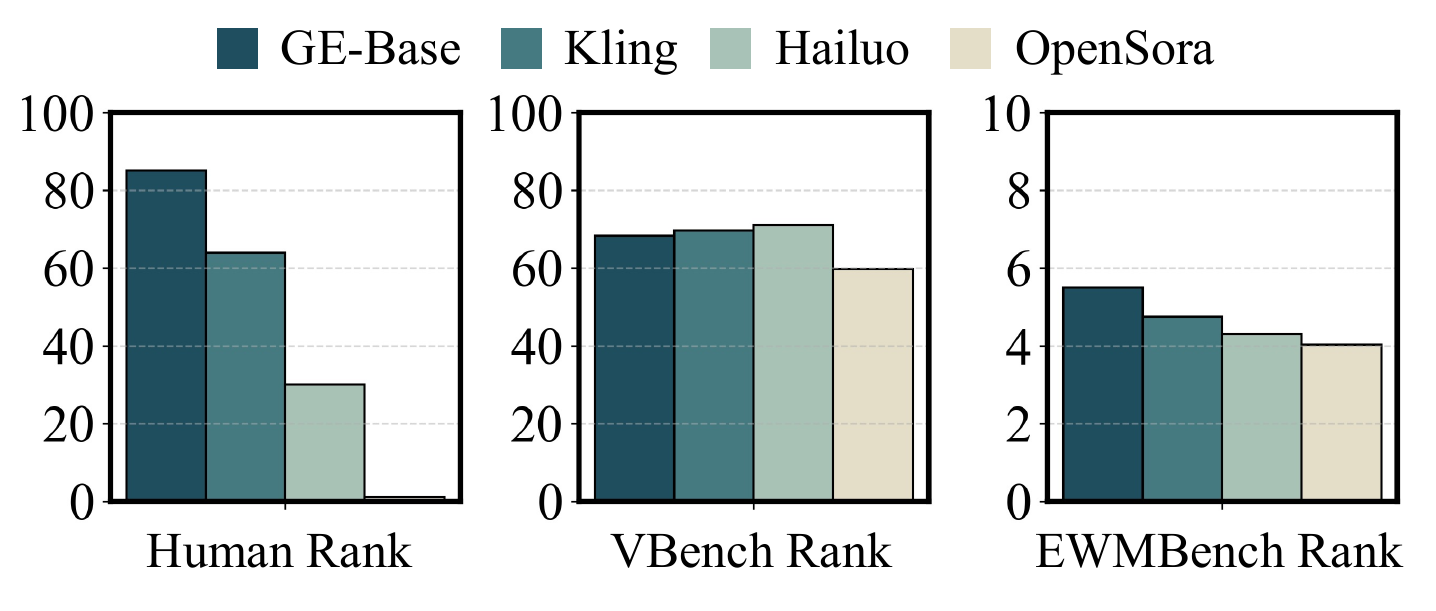}}
        \caption{\textbf{Consistency and Validity Analysis of Evaluation Metrics.} We compare human preference with our proposed EWMBench and the general video benchmark VBench to assess the consistency and reliability of automated evaluation metrics across different video world models.
        }
    \label{fig:exp-benchrank-left}
\end{minipage}
\hfill
\begin{minipage}[b]{0.4\textwidth}
\footnotesize
\centering
\setlength{\tabcolsep}{4pt}
\begin{tabular}{lccccc}
\toprule
\textbf{Model} & BLEU & CLIP & DYN & Div. & PSNR \\
\midrule
LTX     & 0.33 & 90.8 & 0.78 & 0.011 & 19.9 \\
COSMOS  & 0.31 & 90.2 & 0.85 & 0.010 & 20.7 \\
\midrule
\textbf{Model} & SA & Log. & TA & Scn. &  \\
\midrule
LTX     & 0.94 & 0.97 & 0.98 & 0.90 &  \\
COSMOS  & 0.87 & 0.97 & 0.97 & 0.91 &  \\
\bottomrule
\end{tabular}
\vspace{1.5mm}
\captionof{table}{
\textbf{GE-Sim Evaluation on EWMBench.}
Comparison of LTX-Video and COSMOS2 as base models for action-conditioned video generation.
Metrics evaluate spatial, temporal, and semantic alignment with ground-truth control trajectories.}
\label{tab:ac-bench}
\end{minipage}
\end{figure}

In addition to instruction-conditioned evaluation, we further assess the fidelity and reliability of our video-based simulator in an action-conditioned setting across two base models. Given ground-truth action trajectories, the simulator generates visual predictions conditioned solely on these control sequences. Under the EWMBench framework (Table~\ref{tab:ac-bench}), GE-Sim consistently demonstrates high spatial accuracy, precise trajectory execution, and strong semantic coherence. The limited visual diversity under fixed action inputs indicates accurate action-to-video correspondence and robust alignment with control dynamics. While GE-Sim built on COSMOS2 and LTX-Video achieves comparable overall performance, the COSMOS2-based variant excels in dynamic consistency, reflecting superior embodied video generation fidelity and temporal coherence.

\vspace{-2mm}

These results highlight that, beyond its efficiency, low cost, and ability to generalize across diverse environments, our video-based simulator provides a reliable and semantically consistent platform for action-conditioned evaluation in robotic manipulation.

\subsection{Metric-Human Consistency}
To validate the reliability and task relevance of our proposed EWMBench, we conduct a comparative analysis against human preference ratings and the general video benchmark VBench. We collect human annotations on videos generated by four representative models, GE-Base, Kling-1.6~\citep{Kuaishou2025Kling}, Hailuo I2V-01-live~\citep{hailuo}, and OpenSora-2.0~\citep{opensora}, using a ranking protocol where annotators assign ordinal scores based on perceived overall quality. Rankings are aggregated across annotators and samples, with multiple review rounds to ensure annotation consistency. As shown in Figure~\ref{fig:exp-benchrank-left}, empirical results demonstrate that EWMBench rankings exhibit strong concordance with human judgments, effectively capturing dimensions of temporal alignment, semantic fidelity, and visual coherence. In contrast, VBench exhibits misalignment, particularly in scenarios demanding embodied consistency and goal-conditioned reasoning. These results confirm that EWMBench provides a more faithful and task-grounded assessment of video-based world models in robotic manipulation.

\section{Related Works}
\label{sec:related}

\noindent \textbf{World Models for Robotic Manipulation.} The concept of world models as internal predictive representations for perception, planning, and control has long played a central role in robotics~\citep{sutton1981adaptive,chatila1985position}. Early approaches relied on analytical modeling and system identification~\citep{murray2017mathematical}, requiring task-specific engineering and limited generalizability. The introduction of neural world models~\citep{ha2018world} enabled learning compact representations of dynamics directly from sensory inputs. These models have since evolved to operate in both pixel space~\citep{finn2016unsupervised,ebert2018visual} and learned latent spaces~\citep{hafner2019learning,wu2023daydreamer,hu2024learning}, with applications in control and planning. However, most prior efforts remain task-specific, constrained by limited interaction data. Recent advances propose general video-based world models trained on large-scale datasets~\citep{bruce2024genie,agarwal2025cosmos,russell2025gaia,jang2025dreamgen}, yet these primarily focus on visual synthesis and do not support closed-loop robotic control. In contrast, our work develops a unified framework that integrates video-based world modeling with an action decoding module (GE-Act) and a closed-loop simulator (GE-Sim), enabling direct application in real-world robotic manipulation.

\noindent \textbf{Video Generative Models for Robot Learning.} Progress in video generation has led to powerful models capable of synthesizing high-quality videos from text or image prompts~\citep{sora,ho2022video,blattmann2023stable}. While these models achieve impressive visual quality~\citep{sora,yang2024cogvideox,blattmann2023stable}, their application to robotics remains limited by the lack of action conditioning, temporal coherence, and multi-view reasoning. Robotic manipulation requires models that can predict future states conditioned on action instruction, maintain long-term temporal consistency, and reason over spatially distributed observations. Action-conditioned video models~\citep{bruce2024genie} have shown initial promise, and increasingly sophisticated systems are being developed, including driving~\citep{russell2025gaia} and robotic models~\citep{agarwal2025cosmos}, with camera controllability~\citep{wang2024motionctrl} further enhancing manipulation capabilities. However, existing methods are limited to single-view predictions and often lack comprehensive task understanding. GE-Base addresses these limitations through multi-view synthesis and autoregressive decoding with a memory mechanism, improving spatiotemporal consistency and task relevance.

\noindent \textbf{Vision-Language-Action Models.} Vision-language-action (VLA) models have emerged as a dominant paradigm in instruction-conditioned robotics~\citep{driess2023palm-e,brohan2023rt2,kim2024openvla,black2024pi_0}. These models typically initialize from large-scale vision-language pretraining and are fine-tuned on robot demonstrations to predict action sequences. Although this approach has shown strong performance in diverse tasks, it suffers from inherent limitations. Behavior cloning restricts the agent to imitation without the ability to recover from errors or explore alternative strategies. The absence of explicit world models prevents internal simulation or reasoning over potential outcomes. Moreover, collecting high-quality teleoperation data remains a major bottleneck. Alternative approaches attempt to use VLMs as frozen encoders~\citep{nair2022r3m} or high-level planners~\citep{ahn2022can,huang2023instruct2act}. Our framework takes a different approach by using vision-language inputs to condition a generative world model, enabling predictive reasoning and planning through internal simulation.

\noindent \textbf{Policy Evaluation in Robotics.} Efficient policy evaluation is essential for scaling robot learning. Traditional physics engines such as MuJoCo~\citep{todorov2012mujoco} and Isaac Gym~\citep{makoviychuk2021isaac} provide fast simulation but require extensive manual tuning and still face a gap when transferring to the real world. Real-world evaluations, while more accurate, are slow and resource-intensive~\citep{zhou2025autoeval}. Recent efforts incorporate generative models into simulators~\citep{Genesis,nasiriany2024robocasa}, offering new possibilities for efficient and scalable evaluation. However, many of these approaches are limited to simplified settings or restricted observation modalities. GE-Sim addresses these challenges by embedding robotic models within a generative loop that supports long-horizon manipulation across multiple views and includes both successful and failure-mode trajectories to improve robustness and reliability.

\noindent \textbf{Evaluation of Embodied World Models.} Assessing the quality of embodied world models requires metrics that reflect performance in realistic manipulation scenarios. Traditional video generation metrics such as MSE or FVD do not correlate well with real-world task success. Recent benchmarks introduce structured evaluation protocols~\citep{huang2024vbench,huang2024vbench++} with broader metric coverage, but many still emphasize visual realism over task relevance. Specialized frameworks such as PhyGenBench~\citep{phygenbench} and T2V-CompBench~\citep{sun2024t2v} assess physical understanding and compositionality, respectively, but lack alignment with control objectives. Our EWMBench addresses this gap by providing a comprehensive evaluation suite focused on visual fidelity, motion consistency, semantic alignment, and action-conditioned controllability~\citep{yue2025ewmbench}. It is specifically designed to assess the capabilities of video-based world models in the context of embodied robotics.

\section{Limitations}
In this work, we present a systematic investigation into world models for real-world robotic manipulation, addressing core challenges in visuomotor representation, policy learning, and embodied evaluation. While our Genie Envisioner framework lays a foundational path toward scalable and generalizable robotic intelligence, several limitations remain:

\begin{itemize}
\item \emph{Data Coverage and Source Diversity.} Although we conduct cross-embodiment transfer experiments, our training relies exclusively on the AgiBot-World-Beta dataset—a large-scale yet single-platform real-world corpus. No internet-scale or simulation-based data sources are incorporated, limiting the diversity of embodiment types, sensor modalities, and scene configurations encountered during pretraining. While Genie Envisioner demonstrates promising generalization via few-shot adaptation, its robustness across heterogeneous sources and low-resource domains remains underexplored. Future extensions incorporating large-scale simulated or web-derived demonstrations will be critical for further expanding transfer capabilities.

\item \emph{Embodiment Scope and Dexterity.} The current study is confined to upper-body tabletop manipulation using parallel-jaw grippers. More complex embodiment settings, including dexterous hand coordination and full-body locomotion, are not addressed. These capabilities are crucial for real-world general-purpose robotics and warrant further integration into the Genie Envisioner framework to support fine-grained, multi-contact interactions and whole-body behaviors.

\item \emph{Evaluation Methodology.} While our EWMBench provides a structured evaluation of visual fidelity, action consistency, and language grounding, it still relies on proxy metrics and partial human validation. Fully automated and reliable assessment of task success—particularly under diverse failure modes and ambiguous semantics—remains an open challenge. Building scalable evaluation protocols that align closely with human judgment will be essential for robust benchmarking and safe deployment in real-world scenarios.
\end{itemize}

While Genie Envisioner is not yet a complete solution, it represents a meaningful step toward Genie, embodied AI systems with the potential for AGI-level manipulation capabilities.

\section{Conclusion}
In this work, we introduce Genie Envisioner, a unified and scalable platform for dual-arm robotic manipulation, leveraging high-fidelity video generation. At its core, GE-Base provides a robust foundation, capturing the spatiotemporal and semantic dynamics of robotic interactions for instruction-aligned video synthesis. The integration of GE-Act enables high-precision task execution, demonstrating not only strong performance across diverse in-domain tasks but also exceptional cross-embodiment generalization. Through minimal adaptation, GE-Act successfully transfers to novel robotic platforms and excels in complex tasks such as cloth folding and box packing. GE-Sim enhances the framework further by supporting closed-loop simulation, allowing for continuous policy refinement. EWMBench provides a comprehensive evaluation suite, ensuring robust assessment across visual realism, semantic alignment, and policy consistency. Extensive real-world evaluations confirm the superiority of GE-Base, GE-Act, and GE-Sim, establishing Genie Envisioner as a powerful foundation for building general-purpose, instruction-driven embodied intelligence.

\section*{Acknowledgment}
We gratefully acknowledge the foundational contributions of prior works, including EnerVerse~\citep{huang2025enerverse}, EnerVerse-AC~\citep{jiang2025enerverse}, and EWMBENCH~\citep{yue2025ewmbench}, which provided the inspiration and foundation for this research. We appreciate the AgiBot Genie Team for their invaluable contributions to data collection, real-world evaluation, and the provision of both robotic hardware and software support throughout this project.

{
\small
\bibliography{ref}
}

\end{document}